\pdfoutput=1

\documentclass[11pt]{article}

\usepackage[preprint]{acl}
\usepackage{times}
\usepackage{latexsym}
\usepackage[T1]{fontenc}
\usepackage[utf8]{inputenc}
\usepackage{microtype}
\usepackage{inconsolata}

\usepackage{amsmath}
\usepackage{amssymb}
\usepackage{booktabs}
\usepackage{graphicx}
\usepackage{subcaption}
\usepackage{adjustbox}
\usepackage{multirow}

\usepackage{bm}
\usepackage{comment}
\newcommand\blfootnote[1]{
  \begingroup
  \renewcommand\thefootnote{}\footnote{#1}
  \addtocounter{footnote}{-1}
  \endgroup
}

\title{Quantifying Lexical Semantic Shift via Unbalanced Optimal Transport}

\author{
  {\bf Ryo Kishino}${}^{1}$ \qquad
  {\bf Hiroaki Yamagiwa}${}^{1}$ \qquad\\
  {\bf Ryo Nagata}${}^{2,5}$ \qquad
  {\bf Sho Yokoi}${}^{3,4,5}$ \qquad
  {\bf Hidetoshi Shimodaira}${}^{1,5}$ \\
  ${}^1\,$Kyoto University \quad ${}^2\,$Konan University \quad ${}^3\,$NINJAL \quad  ${}^4\,$Tohoku University \quad ${}^5\,$RIKEN\\
  \texttt{kishino.ryo.32s@st.kyoto-u.ac.jp},
  \texttt{h.yamagiwa@i.kyoto-u.ac.jp},\\
  \texttt{nagata-acl2025@ml.hyogo-u.ac.jp},
    \texttt{yokoi@ninjal.ac.jp},
  \texttt{shimo@i.kyoto-u.ac.jp}\\
}
\begin{document}
\maketitle
\begin{abstract}

Lexical semantic change detection aims to identify shifts in word meanings over time. While existing methods using embeddings from a diachronic corpus pair estimate the degree of change for target words, they offer limited insight into changes at the level of individual usage instances. To address this, we apply Unbalanced Optimal Transport (UOT) to sets of contextualized word embeddings, capturing semantic change through the excess and deficit in the alignment between usage instances. In particular, we propose Sense Usage Shift (SUS), a measure that quantifies changes in the usage frequency of a word sense at each usage instance. By leveraging SUS, we demonstrate that several challenges in semantic change detection can be addressed in a unified manner, including quantifying instance-level semantic change and word-level tasks such as measuring the magnitude of semantic change and the broadening or narrowing of meaning.
\blfootnote{Our code is available at \url{https://github.com/ryo-lyo/Semantic-Shift-via-UOT}.}
\end{abstract}

\section{Introduction} \label{sec:introduction}

\begin{figure}[!t]
    \begin{minipage}{\linewidth}
        \centering
        \includegraphics[width=1.0\linewidth]{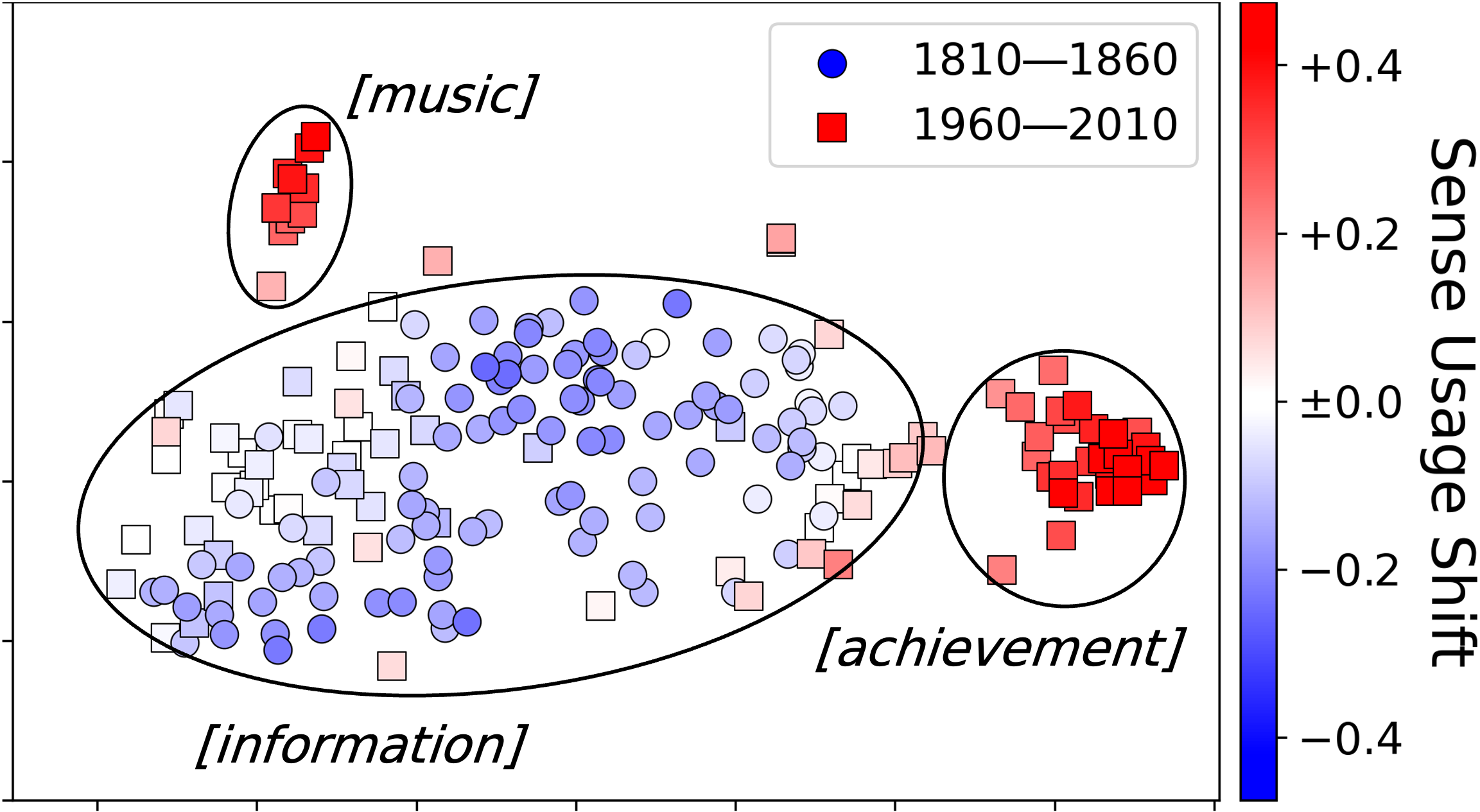}\\
        \subcaption{\textit{record}}\label{fig:record}
    \end{minipage}\\[3mm]
    \begin{minipage}{\linewidth}
        \centering
        \includegraphics[width=1.0\linewidth]{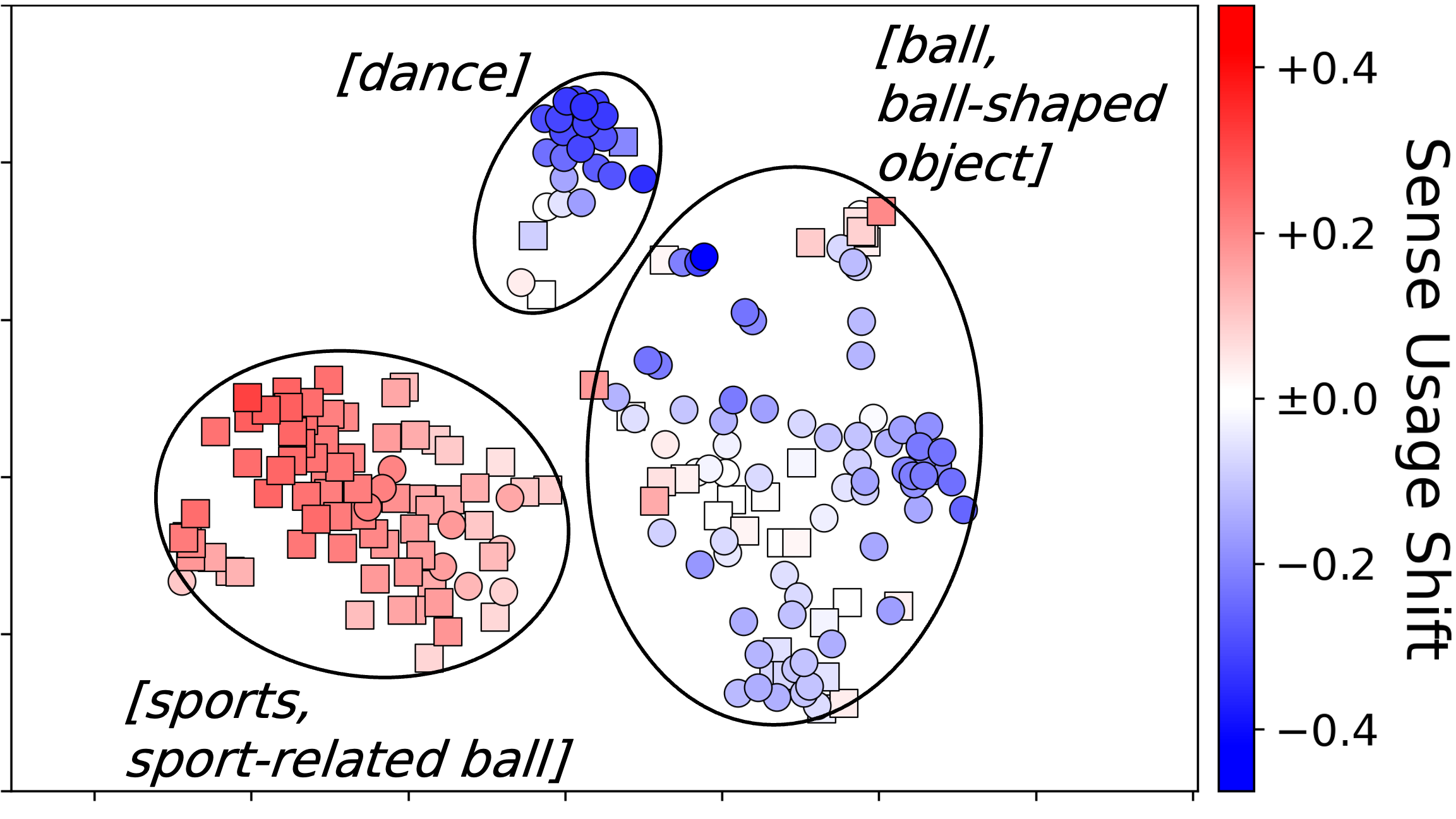}\\
        \subcaption{\textit{ball}}\label{fig:ball}
    \end{minipage}
    \caption{t-SNE visualization of the contextualized embeddings for each usage instance of a target word in the diachronic corpus pair. The color of each instance represents its Sense Usage Shift (SUS). The relative frequency of the target word usage in the senses of the red instances has increased compared to its usage in the other senses, while that of the blue instances has decreased. (a) For the word \textit{record}, instances in the sense of \textit{[music]} and \textit{[achievement]}, whose usage frequencies have increased, exhibit high SUS values. (b) For the word \textit{ball}, instances in the sense of \textit{[dance]}, whose usage frequency has decreased, exhibit low SUS values. For more details, refer to Section~\ref{sec:experiment_visualization}, and for additional word examples, see Appendix~\ref{appx:experiment_visualization}.}
    \label{fig:tsne-plot}
\end{figure}

\begin{table*}[t]
\small
\centering
\begin{tabular}{ccccc}
\toprule
                         & Corpus & Instance                                     & Sense             & SUS \\
\midrule

\multirow{3}{*}{Top~3}    & 1960--2010    &... So did Sire \textbf{\textit{Records}}...                      & undefined (\textit{[music]})       & 0.47 \\
                         & 1960--2010    &... a team with the third-worst \textbf{\textit{record}}...        & \textit{[achievement]} & 0.45                    \\
                         & 1960--2010    &... the AMCU single-season \textbf{\textit{record}}...             & \textit{[achievement]} & 0.45                    \\
\midrule
\multirow{3}{*}{Bottom~3} & 1810--1860    &... interpretations of the Mosaic \textbf{\textit{record}}...      & \textit{[information]}     & -0.23                   \\
                         & 1810--1860    &... the \textbf{\textit{records}} of a professed revelation...     & \textit{[information]}     & -0.24                   \\
                         & 1810--1860    &... the \textbf{\textit{record}} of whose wisdom is included in... & \textit{[information]}     & -0.25                   \\
\bottomrule
\end{tabular}
    \caption{The top 3 and bottom 3 usage instances of the target word \textit{record} based on SUS values. Instances with high SUS values correspond to senses whose usage has increased across the diachronic corpus pair, while instances with low SUS values correspond to senses whose usage has decreased, as discussed in Section~\ref{sec:experiment_visualization}. For other target words, see Table~\ref{tab:appx-sus-tables} in Appendix~\ref{appx:experiment_visualization}. For more details, including `undefined', refer to Appendix~\ref{appx:detail-dwug}.}
    \label{tab:record-sus-example}
\end{table*}

Lexical semantic change detection is the task of identifying words that change their meaning over time, as well as determining which specific senses have disappeared or emerged~\citep{DBLP:journals/csur/PeritiM24}. Recently, methods leveraging word embeddings from a diachronic corpus pair have been studied. Previous approaches detect semantic change by measuring differences in contextualized embeddings of a target word across two corpora~\citep{DBLP:conf/acl/GiulianelliTF20, DBLP:conf/naacl/MontariolMP21, aida-bollegala-2023-unsupervised}.

These methods focus on the entire set of usage instances. From a linguistic point of view, however, it is critical to focus on individual instances as well as on the overall differences. For example, it is important to quantify the extent to which a disappearing or emerging sense of a word loses or acquires its popularity compared to the other senses in order to reveal how and why the word has undergone the semantic change. Unfortunately, there have been almost no studies on this topic.

To address this limitation, we apply Unbalanced Optimal Transport (UOT) between the two sets of contextualized embeddings of a target word obtained from a diachronic corpus pair, focusing on the excess and deficit in the alignment between usage instances. Using this alignment discrepancy, we propose a novel measure called Sense Usage Shift (SUS), which quantifies how much the relative frequency of the word usage in a word sense of each instance has changed across the corpora.

Fig.~\ref{fig:tsne-plot} illustrates that word usage in the word sense of instances with high SUS values has become frequent in the modern corpus, whereas usage in the sense of instances with low SUS values has decreased. Table~\ref{tab:record-sus-example} shows several usage instances with high or low SUS values. Although the previous work~\citep{DBLP:conf/naacl/MontariolMP21} has applied standard Optimal Transport (OT) for detecting semantic change, the balanced alignment fails to capture information about individual instances, as illustrated in Fig.~\ref{fig:OT_UOT_illust} and Fig.~\ref{fig:record-ot-uot-matrix}.

Moreover, by leveraging the SUS values calculated for each usage instance, we address various tasks in semantic change in a unified manner. Specifically, for a given target word, SUS enables (1) the quantification of semantic change at the instance level, (2) the quantification of semantic change at the word level, and (3) the quantification of the extent to which the meaning of a target word has broadened or narrowed. Experiments demonstrate that the proposed SUS-based methods achieve performance comparable to or better than existing approaches for these tasks.

\section{Related Work}\label{sec:related_work}
\subsection{Lexical semantic change detection}

One of the major approaches to lexical semantic change detection is to align two sets of static word embeddings to measure semantic differences between a diachronic corpus pair. This can be done either by assuming linear transformations~\citep{DBLP:conf/www/KulkarniAPS15, hamilton-etal-2016-diachronic} or without the linear assumption~\citep{DBLP:conf/wsdm/YaoSDRX18, aida-etal-2021-comprehensive} between the two embedding spaces. However, static embedding-based methods can only handle word-level semantic change, providing no information on changes at the sense level or the individual instance level.

To address this, methods leveraging contextualized word embeddings for sets of usage instances of a target word, obtained from old and modern corpora using language models such as BERT~\citep{DBLP:conf/naacl/DevlinCLT19}, have been proposed. These methods are broadly categorized into sense-based and form-based approaches~\citep{DBLP:conf/acl/GiulianelliTF20, DBLP:journals/csur/PeritiM24} as follows.

\paragraph{Sense-based approach.} This approach aims to estimate, either directly or indirectly, the frequency of word usage corresponding to specific senses of the target word in two given corpora. The estimated sense counts are then directly used to compute the degree of semantic change. \citet{DBLP:conf/acl/GiulianelliTF20, DBLP:conf/semeval/RotherHE20, DBLP:conf/semeval/KutuzovG20, DBLP:conf/acl-lchange/PeritiFMR22} applied clustering algorithms to two sets of contextualized embeddings of the target word to identify sense counts. Clustering methods such as $K$-means and Affinity Propagation~\citep{doi:10.1126/science.1136800} are commonly used for this purpose. However, $K$-means faces the non-trivial challenge of determining the number of senses, while Affinity Propagation suffers from unstable clustering results~\citep{DBLP:journals/csur/PeritiM24}.

\paragraph{Form-based approach.} This approach detects semantic change by comparing the probability distributions of contextualized word embeddings from the two corpora. \citet{aida-bollegala-2023-unsupervised} and \citet{nagata-etal-2023-variance} assume that embeddings follow a normal distribution and a von Mises-Fisher (vMF) distribution, respectively, and define the degree of semantic change as the difference between the two distributions.

Unlike the sense-based approach, the form-based approach bypasses sense identification, which is likely to result in more stable semantic change detection. As mentioned in Eqs.~(3) and (8) of \citet{nagata-etal-2023-variance}, it is also possible to capture semantic change at the instance level by using the ratio of probability densities in the two corpora. While \citet{nagata-etal-2023-variance} discuss this approach, to the best of our knowledge, prior research has not quantitatively evaluated the detection of semantic change at the instance level.

\subsection{Comparing distributions of embeddings with optimal transport}

OT is a method for measuring the distance between two probability distributions through complete alignment, while UOT allows for excess and deficit in the alignment. In natural language processing, OT is widely used to compute the distance between two documents based on embeddings~\citep{DBLP:conf/icml/KusnerSKW15, DBLP:conf/emnlp/YokoiTASI20}, and UOT is also utilized for this purpose~\citep{DBLP:conf/acml/WangZY0RW20, DBLP:conf/ijcai/ChenLXPMC20, DBLP:conf/acl/SwansonYL20, DBLP:conf/acl/Arase0Y23, DBLP:conf/acl/ZhaoWZW20}.

In semantic change detection, \citet{DBLP:conf/naacl/MontariolMP21} were the first to utilize OT to compute the degree of semantic change for a target word. Their approach involved clustering the contextualized embeddings of a target word obtained from two corpora and applying OT to the sets of centroids corresponding to each sense. Consequently, unlike the proposed method, this approach does not provide information about individual usage instances except through the identified sense clusters. On the other hand, \citet{DBLP:journals/corr/abs-2402-16596} directly applied OT to two sets of contextualized embeddings. However, to the best of our knowledge, no existing studies have utilized UOT for semantic change detection, as proposed in our method.

\section{Background on Optimal Transport}

\subsection{Problem setting}

We aim to identify how the meanings of a target word $w$ change across a diachronic corpus pair. A context containing the target word $w$ is referred to as \textit{a usage instance of} $w$. For simplicity, it is referred to as \textit{an instance}. The sense of $w$ in a usage instance is referred to as \textit{the sense of an instance}. Let the set of $m$ usage instances of the target word $w$ belonging to the old corpus be $\{s^w_i\}_{i=1}^m$. Likewise, let the set of $n$ usage instances of $w$ belonging to the modern corpus be $\{t^w_j\}_{j=1}^n$. Using a pre-trained language model, the contextualized embeddings $\bm{u}^w_i$ and $\bm{v}^w_j \in \mathbb{R}^d$ are calculated for each usage instance $s^w_i$ and $t^w_j$, respectively. Hereafter, the superscript $w$ is omitted for brevity. To detect how the meaning of $w$ has changed between the corpora, we measure the difference between the distributions $\{\bm{u}_i\}_{i=1}^m$ and $\{\bm{v}_j\}_{j=1}^n$.

\subsection{Optimal transport}
For $n \in \mathbb{N}$, let $\bm{1}_n$ denote an $n$-dimensional vector with all elements equal to 1 and $\mathbb{R}_+ = [0, \infty)$ represent the set of non-negative real numbers.

Let \( a_i \in \mathbb{R}_+ \) and \( b_j \in \mathbb{R}_+ \) denote the weights associated with the contextualized word embeddings \( u_i \) and \( v_j \), respectively.
satisfying $\sum_i a_i = \sum_j b_j = 1$. Thus, these weighted embeddings can be regarded as probability distributions. Let $C_{ij}\in \mathbb{R}_+$ denote the transportation cost between the embeddings $\bm{u}_i$ and $\bm{v}_j$, and $T_{ij}\in \mathbb{R}_+$ the amount of transportation. The OT problem, which minimizes the total transportation cost, is formulated as follows:
\begin{align}
\label{eq:OT}
\begin{split}
    \min_{\bm{T} \in \mathbb{R}_+^{m \times n}} &\sum_{i,j} T_{ij} C_{ij} \\
    \mathrm{s.t.~} & \bm{T}\bm{1}_n = \bm{a}, \quad \bm{T}^\top \bm{1}_m = \bm{b},
\end{split}
\end{align}
where $\bm{a} = (a_1, \ldots, a_m)^\top$, $\bm{b} = (b_1, \ldots, b_n)^\top$, and $\bm{T} = (T_{ij})$. The optimization variable $\bm{T}$ is referred to as the transportation matrix, which can be interpreted as providing the alignment between the two sets of the embeddings. Furthermore, the coupling constraints $\bm{T}\bm{1}_n = \bm{a}, \bm{T}^\top\bm{1}_m = \bm{b}$ ensure that there is no excess or deficit in the alignment between the two sets of the embeddings.

\subsection{Limitation of OT in semantic change detection}
\label{sec:OT-limitation}
\begin{figure}[!t]
    \begin{minipage}{\linewidth}
        \centering
        \includegraphics[width=1.0\linewidth]{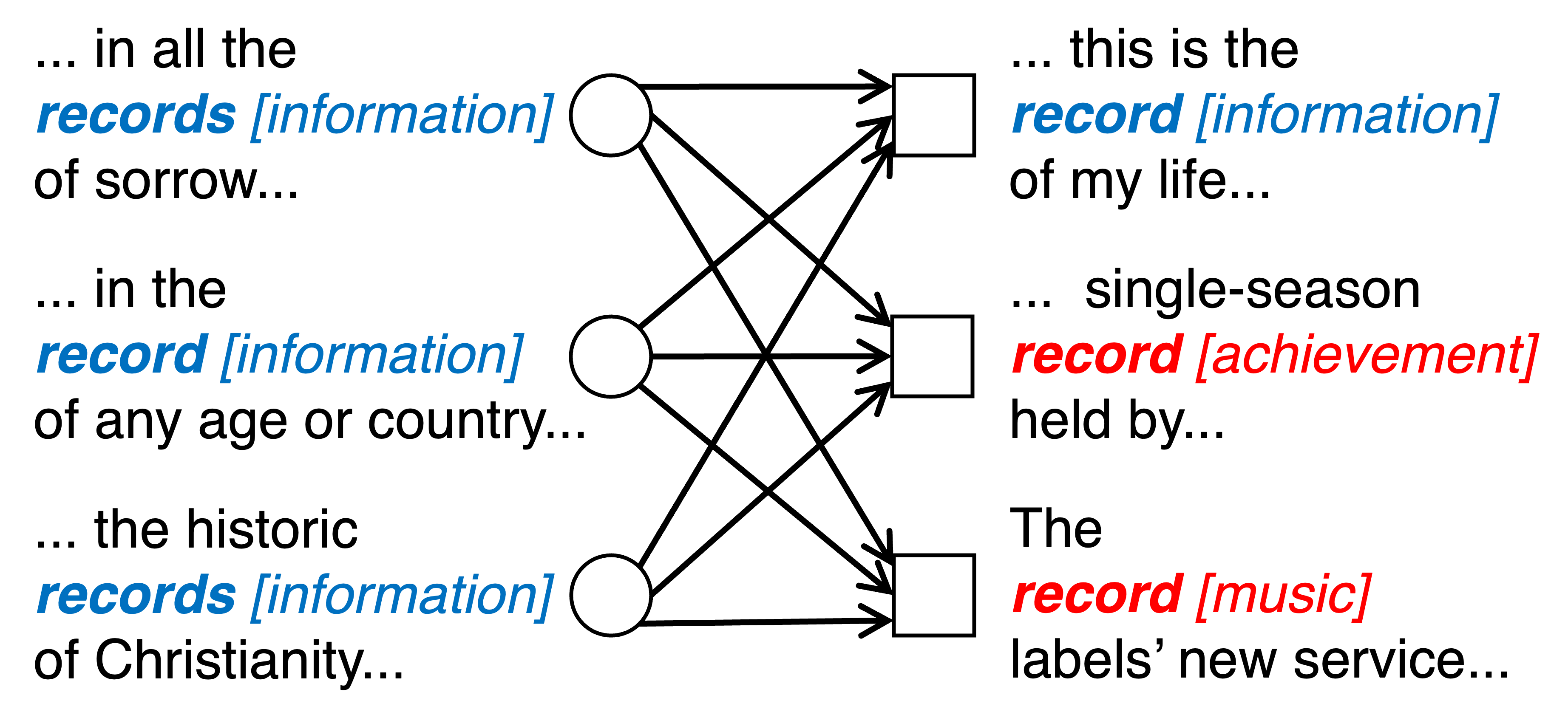}\\
        \subcaption{Balanced alignment in OT}\label{fig:OT}
    \end{minipage}\\[3mm]
    \begin{minipage}{\linewidth}
        \centering
        \includegraphics[width=1.0\linewidth]{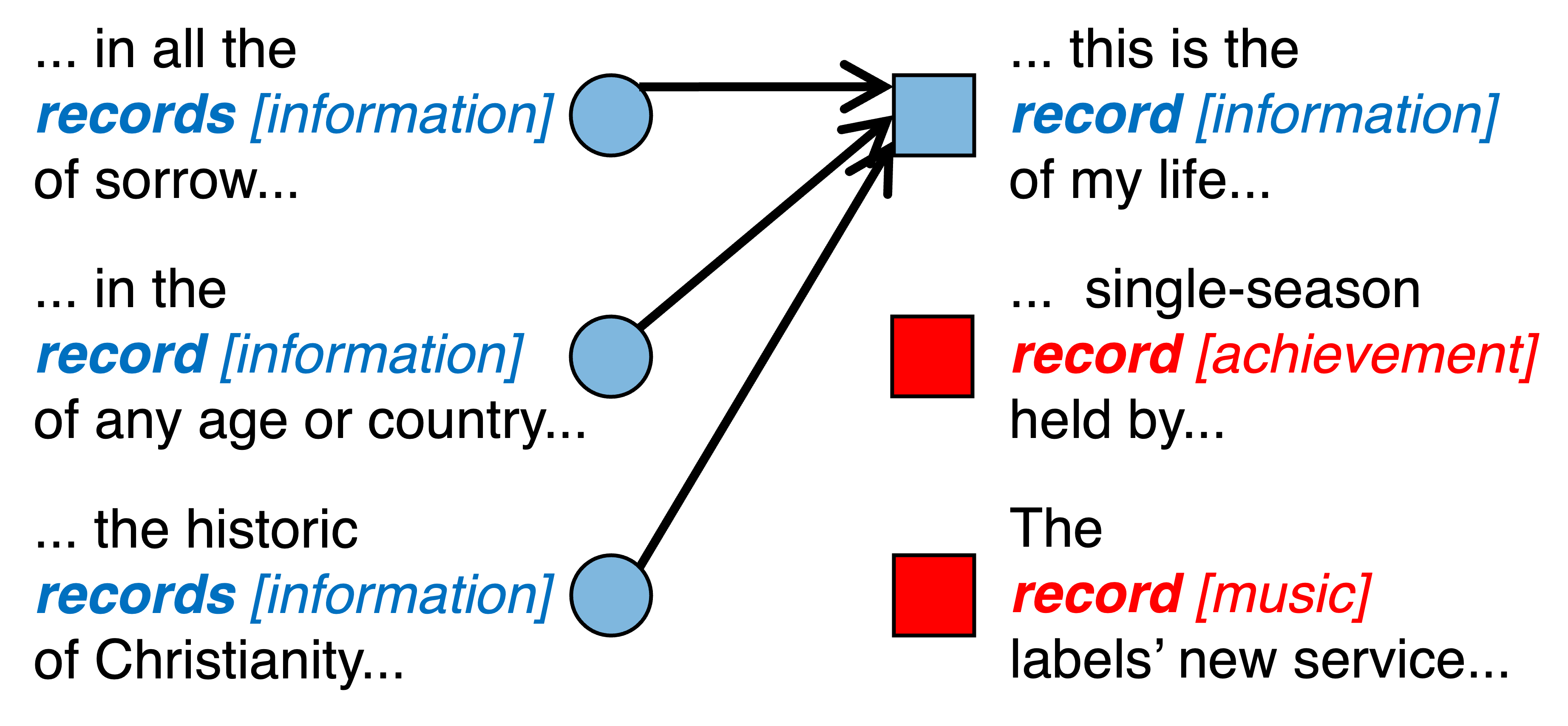}\\
        \subcaption{Unbalanced alignment in UOT}\label{fig:UOT}
    \end{minipage}
    \caption{Illustration of the difference between OT and UOT. Circles and squares represent usage instances of the word \textit{record} in the old and modern corpora, respectively, and the arrows indicate transportation. In the example above, the two bottom squares on the modern side represent instances in the newly emerged senses. (a) OT enforces a balanced alignment between the two sets of instances, which fails to model semantic change, as discussed in Section~\ref{sec:OT-limitation}. (b) UOT allows for excess or deficit in alignment.
    The lack of transported weight from the old side to the two bottom squares on the modern side reflects the increased word usage of these new senses in the modern corpus, as described in Section~\ref{sec:UOT-proposed}. See Fig.~\ref{fig:record-ot-uot-matrix} for a visualization of the transport matrix based on actual numerical experiments.
    }
    \label{fig:OT_UOT_illust}
\end{figure}

\begin{figure}[t]
    \centering
    \includegraphics[width=0.8\linewidth]{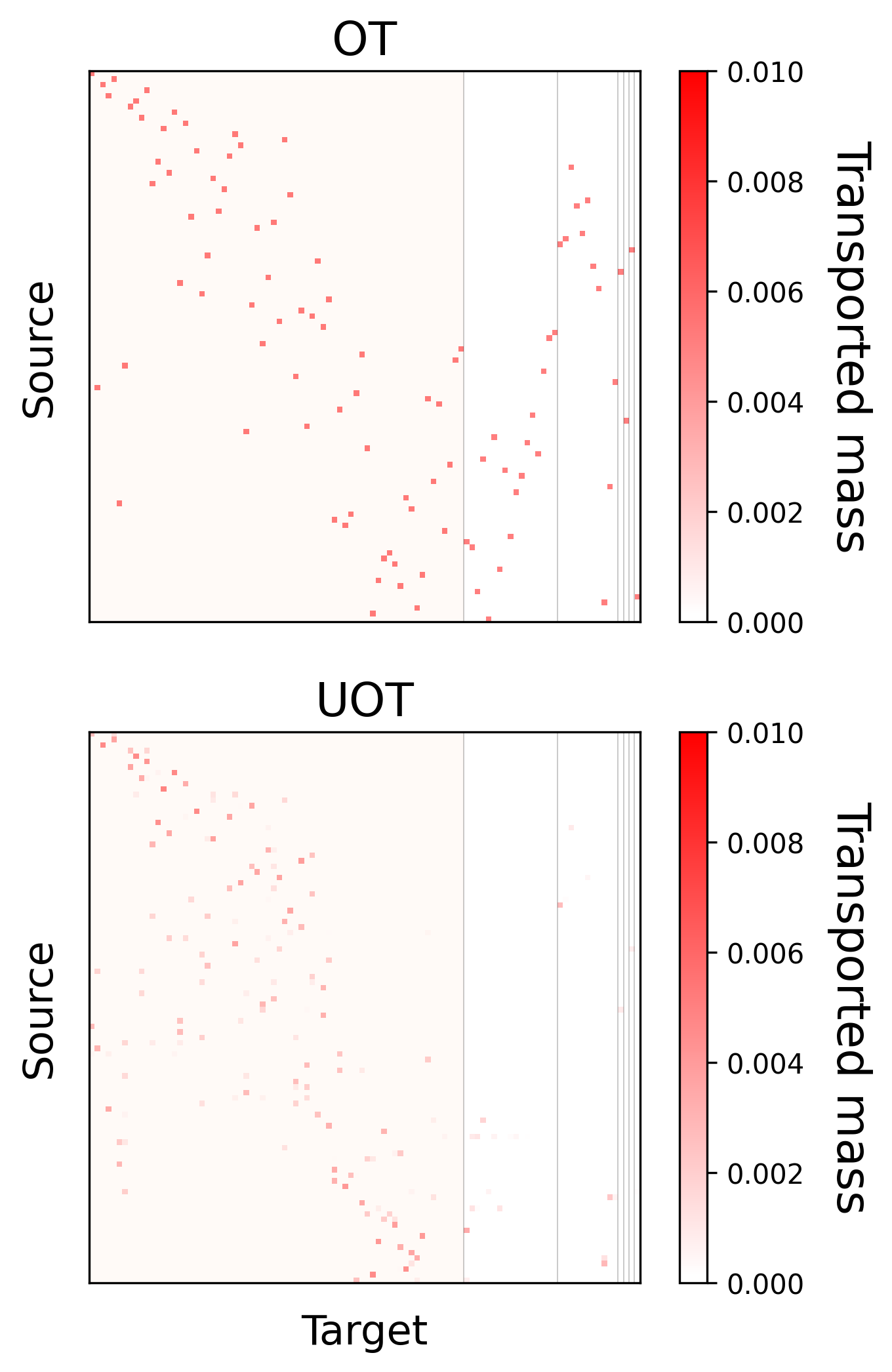}
    \caption{The transportation matrix of OT and UOT for the target word \textit{record}. Red-shaded areas indicate alignments between instances in the same word sense across the two corpora, while white areas indicate alignments between different senses. (Left) OT conducts transportation across different senses. (Right) UOT reduces such transportation by allowing alignment discrepancy. For additional word examples, refer to Fig.~\ref{fig:appx-OT-UOT-matrix} in Appendix~\ref{appx:experiment_visualization}.
    }
    \label{fig:record-ot-uot-matrix}
\end{figure}

\citet{DBLP:journals/corr/abs-2402-16596} utilized the value of the total transportation cost $\sum_{i,j} T_{ij} C_{ij}$ with the optimized $\bm{T}$ as a measure of semantic change at the word level. Standard OT establishes balanced alignment between embeddings (Fig.~\ref{fig:OT}). However, when $w$ undergoes semantic change, substantial shifts can occur between the two corpora, such as the emergence of new senses or the disappearance of existing ones. These shifts pose challenges for OT, as its balanced alignment assumption may not fully account for such large semantic changes. As can be seen from the example of the actual transportation matrix $\bm{T}$ shown in Fig.~\ref{fig:record-ot-uot-matrix}, OT frequently conducts transportation between different senses. As a result, while OT can capture overall semantic change at the word level, it may not fully capture semantic changes in individual usage instances.

\section{Proposed Method}
\subsection{Modeling semantic change with UOT}
\label{sec:UOT-proposed}

The balanced alignment in OT, enforced by the coupling constraints in \eqref{eq:OT}, does not always reflect real-world semantic changes, especially in cases where new senses emerge or existing senses disappear. To address this issue, we consider the following problem, which relaxes these constraints in \eqref{eq:OT}:
\begin{align}
\label{eq:UOT}
    \begin{split}
    \min_{\bm T\in\mathbb R_+^{m\times n}} & \sum_{i,j} T_{ij}C_{ij}+ \lambda_1 D_1(\bm T\bm 1_n, \bm a) \\
    &+ \lambda_2 D_2(\bm T^\top \bm 1_m, \bm b).
    \end{split}
\end{align}
Here, $D_1:\mathbb{R}^m \times \mathbb{R}^m \to \mathbb{R}$ and $D_2:\mathbb{R}^n \times \mathbb{R}^n \to \mathbb{R}$ are penalty functions, and $\lambda_1, \lambda_2$ are hyperparameters that control the degree of penalization. The problem~\eqref{eq:UOT} is referred to as Unbalanced Optimal Transport (UOT). UOT allows for excess or deficit in the alignment by incurring a certain cost.
In our experiments, we assigned uniform weights to each embedding and employed the L2 error as the penalty function. For details, see Section~\ref{sec:experiment_preliminary}.

We focus on the alignment discrepancy represented by $\bm{T}$ in \eqref{eq:UOT} (Fig.~\ref{fig:UOT}). As can be seen from the example in Fig.~\ref{fig:record-ot-uot-matrix}, the transportation in UOT tends to be restricted within the same sense, allowing the alignment excess or deficit to reflect changes in the usage frequency of each sense. With respect to $s_i$, if less weight is transported to the modern corpus than the original weight $a_i$, this indicates that instances in the word sense of $s_i$ are relatively scarce in the modern corpus compared to other $w$ senses. In other words, the usage of $w$ in the word sense of $s_i$ has decreased. Similarly, for $t_j$, if less weight is received from the old corpus than the original weight $b_j$, this suggests that instances in the word sense of $t_j$ are relatively scarce in the old corpus. In other words, the usage of $w$ in the word sense of $t_j$ has increased.

\subsection{Sense usage shift}
\label{sec:sus-definition}

To quantitatively measure semantic change via the excess or deficit in the alignment from UOT, we define Sense Usage Shift (SUS) as a measure of how the frequency of word usage in the word sense of each usage instance $s_i$ or $t_j$ has changed relative to the usage frequencies in the other senses. SUS is defined as follows:
\begin{align}
    \mathrm{SUS}(s_i) &=- (a_i - \sum_j T_{ij}) / a_i, \label{eq:sus1} \\
     \mathrm{SUS}(t_j) &= (b_j - \sum_i T_{ij}) / b_j . \label{eq:sus2}
\end{align}
In other words, SUS represents the excess or deficit in the alignment normalized by the original weight. From the discussion in Section~\ref{sec:UOT-proposed}, a higher SUS value for a usage instance indicates more frequent usage of the target word in the word sense of that instance in the modern corpus. Conversely, a lower SUS value suggests less frequent usage of the target word in that sense. In particular, in \eqref{eq:sus1}, the sign is inverted to ensure that a positive SUS value for an instance corresponds to increased usage in its sense.

\subsection{Understanding semantic change using SUS} 
\label{sec:experiment_visualization}

\begin{figure}[t]
    \begin{minipage}{\linewidth}
        \centering
        \includegraphics[width=1.0\linewidth]{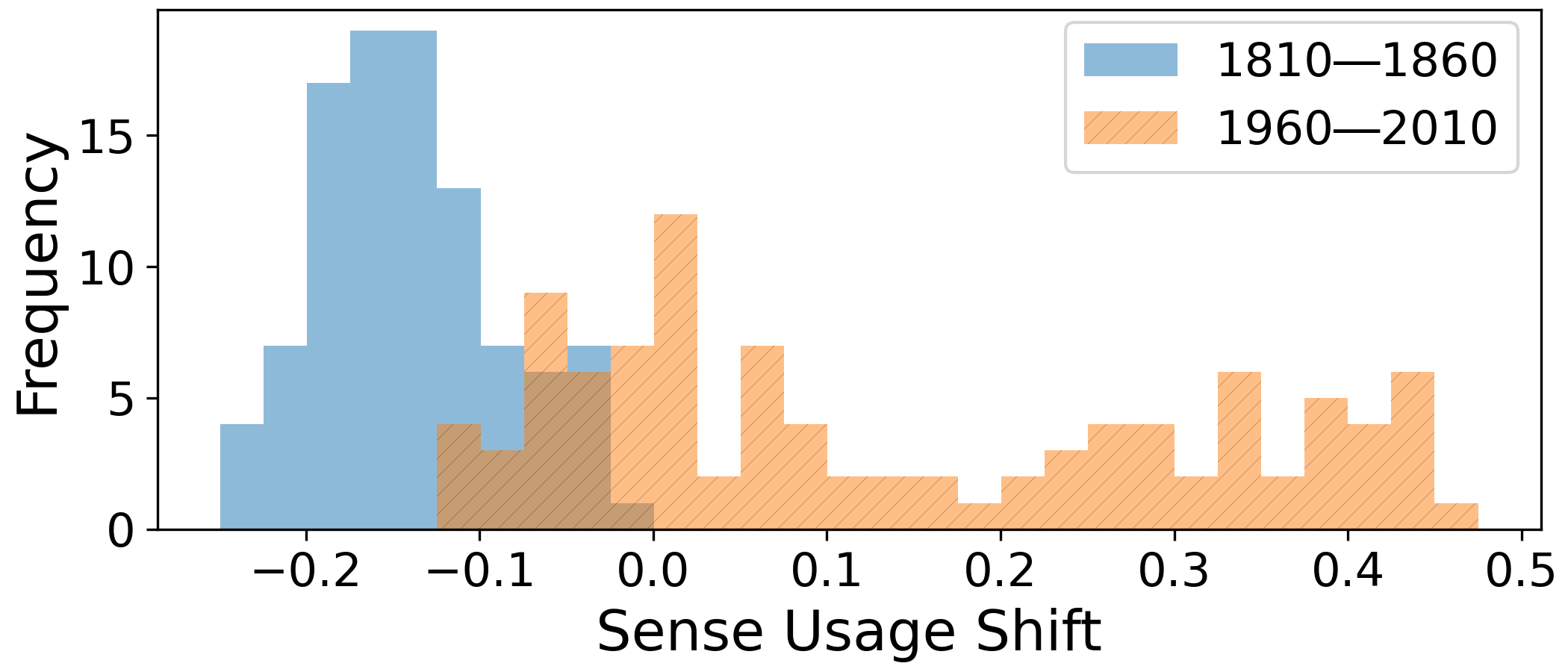}
        \subcaption{\textit{record}}\label{fig:record-sus-hist} 
    \end{minipage}\\[3mm]
    \begin{minipage}{\linewidth}
        \centering 
        \includegraphics[width=1.0\linewidth]{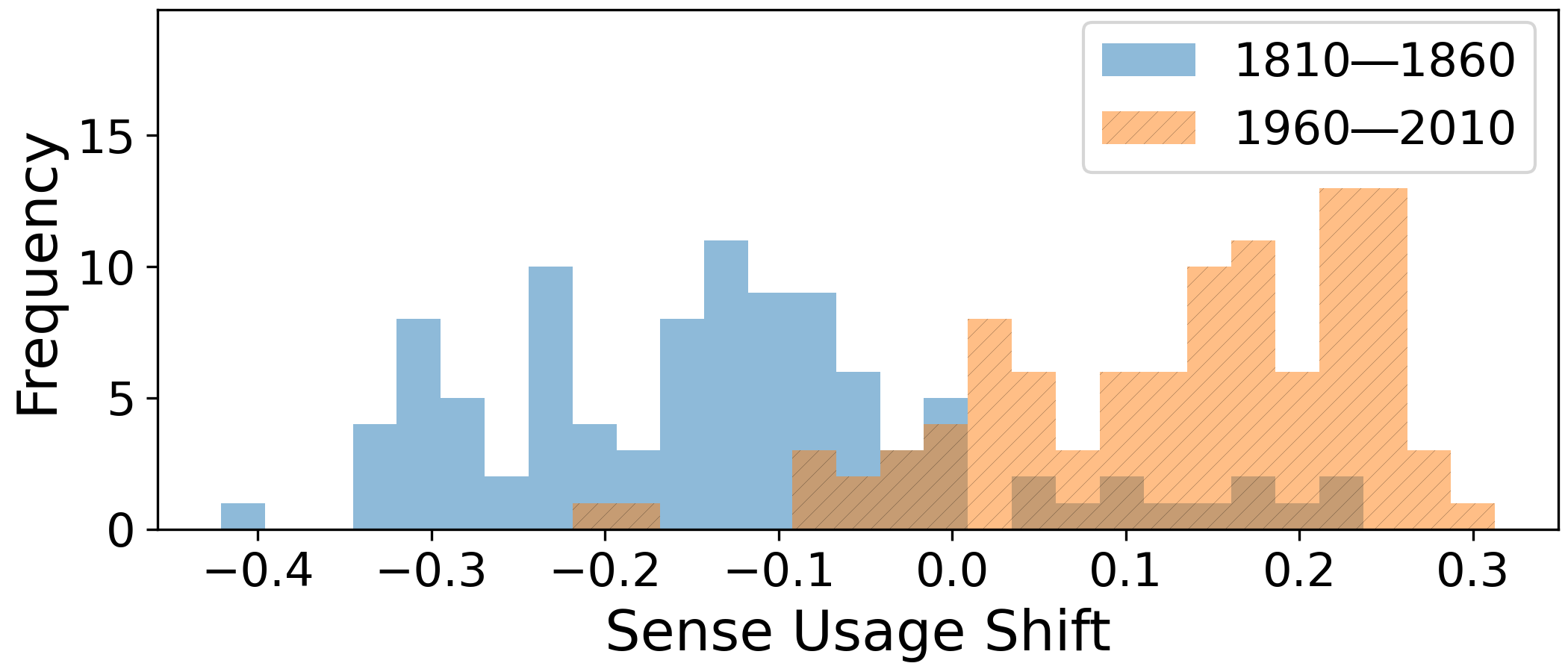}
        \subcaption{\textit{ball}}\label{fig:ball-sus-hist}
    \end{minipage}
    \caption{The distribution of SUS values for usage instances of the target words. (a) For the target word \textit{record}, the variance of SUS values is larger in the modern corpus. This indicates that some usage instances associated with senses in the old corpus also appear in the modern corpus. Therefore, the SUS distribution suggests that \textit{record} is used with a broadened meaning in the modern corpus. (b) For the target word \textit{ball}, conversely, the variance of SUS values is larger in the old corpus, indicating that \textit{ball} is used with a narrowed meaning in the modern corpus.}
    \label{fig:sus-hist}
\end{figure}

To gain deeper insights into SUS defined in \eqref{eq:sus1} and \eqref{eq:sus2}, we visualize SUS values. We focus on the target words \textit{record} and \textit{ball}, each with 100 usage instances in the Diachronic Word Usage Graph (DWUG) dataset~\citep{schlechtweg-etal-2021-dwug, DBLP:conf/emnlp/SchlechtwegCNAW24}. In this dataset, the meaning of \textit{record} has broadened, while that of \textit{ball} has narrowed across the two corpora. Using a pre-trained language model, we compute the contextualized word embeddings $ \{\bm{u}_i\}_{i=1}^{100} $ and $\{\bm{v}_j\}_{j=1}^{100}$. For more details on the experimental setup, refer to Section~\ref{sec:experiment_preliminary}.

\paragraph{Visualizing semantic change with SUS.}
Fig.~\ref{fig:tsne-plot} presents a two-dimensional visualization of the contextualized embeddings of the target words \textit{record} and \textit{ball}, generated using t-SNE\footnote{In the t-SNE configuration, we set the perplexity to 30, the default value in the scikit-learn implementation.}. In this plot, the clusters\footnote{In Fig.~\ref{fig:tsne-plot}, the clusters and their sense labels were manually examined by the authors to represent the sense of the instances. While not all instances were explicitly or strictly annotated, efforts were made to maintain consistency within each cluster. The same procedure was applied to Fig.~\ref{fig:appx-sus-ldr} in Appendix~\ref{appx:experiment_visualization}.} represent the senses of the target words. We analyzed the usage instances within each cluster and identified the sense associated with them. Additional visualization examples for other target words are provided in Fig.~\ref{fig:appx-sus-all}, and usage instances with high or low SUS values are listed in Table~\ref{tab:appx-sus-tables} in Appendix~\ref{appx:experiment_visualization}.

By examining the clusters in these figures, it can be observed that usage instances associated with senses whose frequency has increased or newly emerged have high SUS values, while those associated with senses whose frequency has decreased or disappeared have low SUS values.

\paragraph{Distribution of SUS.}
Furthermore, the distribution of SUS values allows us to interpret whether the meaning of a word has broadened or narrowed. Fig.~\ref{fig:sus-hist} illustrates the distribution of SUS values for usage instances of the target words.  
If the SUS values are regarded as surrogate values for embeddings, a larger variance in SUS values indicates a broader semantic scope.
Accordingly, we infer that the meaning of \textit{record} has broadened, whereas the meaning of \textit{ball} has narrowed.

\subsection{Log-density ratio}
\label{sec:ldr}
\begin{figure}[t]
\begin{minipage}{\linewidth}
    \centering
    \includegraphics[width=1.0\linewidth]{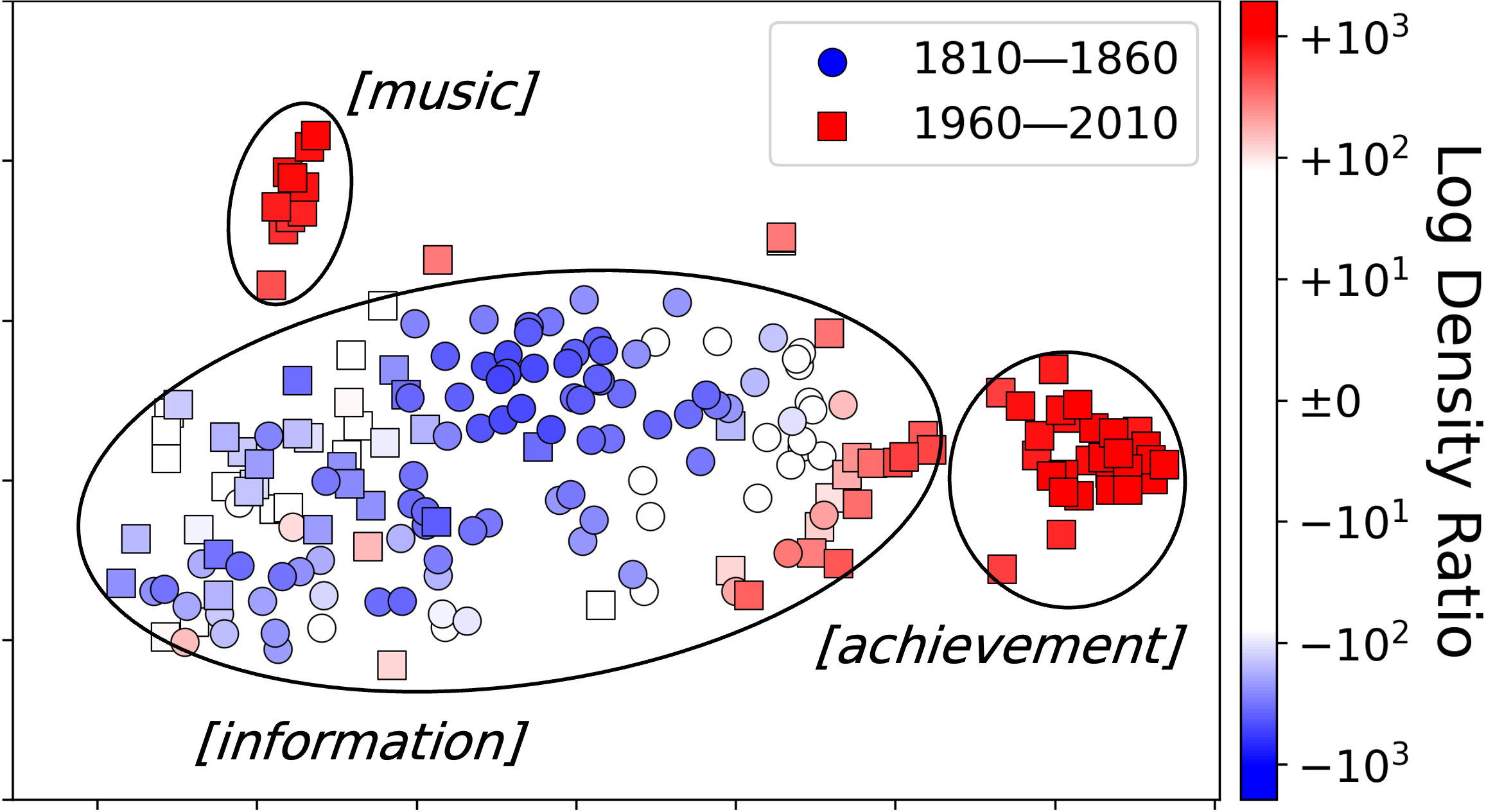}\\
    \subcaption{LDR}
    \label{fig:ldr-tsne-plot}
\end{minipage}\\ 
\begin{minipage}{\linewidth}
    \centering
    \includegraphics[width=0.9\linewidth]{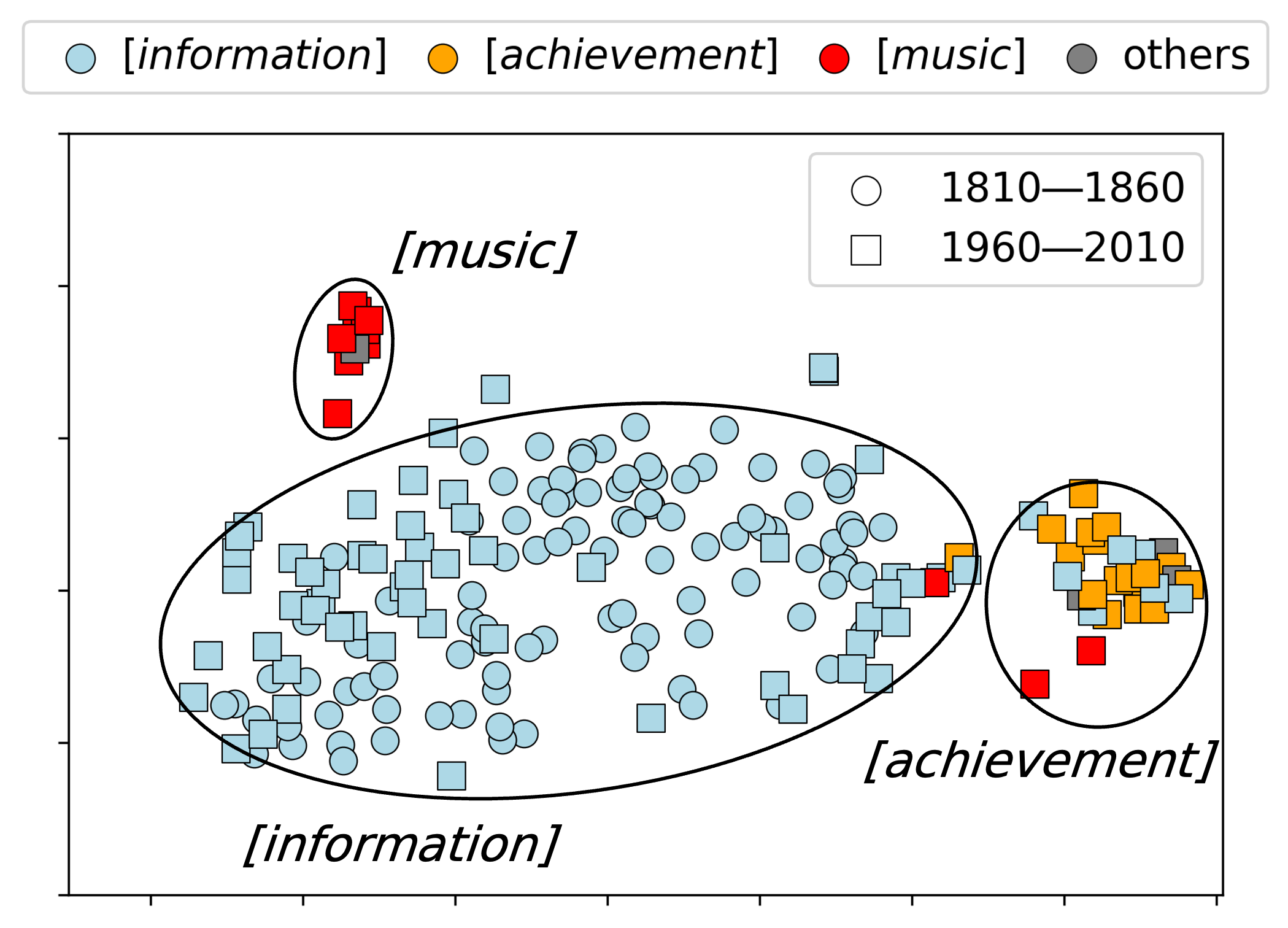}\\
    \subcaption{DWUG}
    \label{fig:appx-record-tsne-dwug}
\end{minipage}
\caption{
    Re-rendering of the t-SNE visualization for the word \textit{record} in Fig.~\ref{fig:record}. The color of each instance represents (a) the Log-Density Ratio (LDR) and (b) the gold sense in the DWUG dataset.
    For details and additional word examples with LDR coloring, refer to Fig.~\ref{fig:appx-ldr-all} in Appendix~\ref{appx:experiment_visualization}. 
    For details on the gold sense, see Section~\ref{subsec:sfd}.
    }
\end{figure}

\paragraph{Definition and computation.}
As a baseline for SUS in the form-based approach, we consider the log-density ratio (LDR). LDR is defined as the logarithm of the ratio of two probability density functions, which are parametrically estimated from the embeddings in the old and modern corpora. The estimation of density ratios has been extensively studied in statistics and machine learning~\citep{sugiyama2012density}. 

In the context of semantic change detection, LDR is mentioned in Eq.~(8) of \citet{nagata-etal-2023-variance}.
Note that they use the term log-likelihood ratio to refer to the LDR at a single instance, although it typically refers to comparisons between densities over the entire dataset.

In our setup, we compute LDR as follows. First, following \citet{nagata-etal-2023-variance}, we assume that the normalized old and modern embeddings follow a distinct von Mises-Fisher (vMF) distribution\footnote{The probability density function of the vMF distribution with mean direction parameter $\bm\mu$ and concentration parameter $\kappa$ is given by
$p(\bm x \mid \bm\mu, \kappa) \propto \exp(\kappa \bm\mu^\top \bm x)$, where $\|\bm x\| = 1$.
}, $p_S$ and $p_T$. Then, we perform maximum likelihood estimation of the parameters using an approximate closed-form solution, as described in Appendix~\ref{appx:ldr}. Using the estimated parameters, we compute 
\begin{align*}
    \mathrm{LDR}(s_i) &= \log \frac{p_T(\tilde{\bm{u}}_i)}{p_S(\tilde{\bm{u}}_i)},\\
    \mathrm{LDR}(t_j) &= \log \frac{p_T(\tilde{\bm{v}}_j)}{p_S(\tilde{\bm{v}}_j)},
\end{align*}
where $\tilde{\bm u}_i = \bm u_i / \|\bm u_i\|$ and $\tilde{\bm v}_j = \bm v_j / \|\bm v_j\|$.

\paragraph{Comparison between SUS and LDR.}
A visual comparison of Fig.~\ref{fig:tsne-plot} and Fig.~\ref{fig:ldr-tsne-plot} suggests that both SUS and LDR roughly reflect sense clusters appropriately. However, referring to the gold senses\footnote{In the dataset used in the experiments, word senses are defined automatically based on human annotations. See Section~\ref{subsec:sfd} for further details.} provided by the DWUG dataset in Fig.~\ref{fig:appx-record-tsne-dwug}, it is evident that SUS reflects the clusters more accurately than LDR. In particular, in the rightmost part of the cluster corresponding to the sense \textit{[information]}, there is a red region where circles and boxes are mixed. 
LDR tends to separate this region as a distinct cluster, although this separation does not match the gold senses. SUS, on the other hand, better preserves the overall cluster structure.
For preprocessing on color configuration, see
Appendix~\ref{appx:experiment_visualization}.

LDR values often fluctuate substantially, making their estimation in high-dimensional spaces unstable. SUS offers a non-parametric alternative that bypasses this instability via UOT.

A detailed comparison between SUS and LDR in semantic change detection tasks will be presented later in Sections~\ref{sec:experiment_instance-level} and \ref{sec:experiment_word-level}.

\subsection{Aggregating SUS to quantify word-level semantic change}
\label{sec:sus-based-word-level}

SUS quantifies semantic change at the level of individual usage instances. By aggregating SUS across all usage instances of a target word, we quantify the degree of word-level semantic change.

We consider two aspects of word-level semantic change. The first is unsigned change, which focuses on the magnitude of semantic shift, regardless of its direction. The second is signed change in the scope of word meaning, which captures the direction of the change. We propose metrics to quantify the degree of each change, which are evaluated in Section~\ref{sec:experiment_word-level}. Due to space constraints, only one metric for each aspect of change is presented in the main text, while additional metrics are provided in Appendix~\ref{appx:experiment_word-level}.

\paragraph{Unsigned change.}
We define the metric $f_\mathrm{SUS}$ to quantify the magnitude of semantic shift. It is computed as the absolute difference between the mean values of the SUS distributions for the two corpora. This metric is given by:
\begin{align*}
    f_{\mathrm{SUS}}(w) =  \left| \frac{1}{m}\sum_{i=1}^m \mathrm{SUS}(s_i) - \frac{1}{n}\sum_{j=1}^n \mathrm{SUS}(t_j) \right|.
\end{align*}

\paragraph{Signed change.}
For a target word, let the variances of the SUS distributions in the old and modern corpora be $ V_S = \mathrm{Var}(\{\mathrm{SUS}(s_i)\}_{i=1}^m)$ and $ V_T = \mathrm{Var}(\{\mathrm{SUS}(t_j)\}_{j=1}^n)$, respectively. As discussed in Section~\ref{sec:experiment_visualization}, the difference in SUS variances between the two corpora reflects the broadening or narrowing of word meaning. Based on this observation, we define the metric $g_{\mathrm{SUS}}$, which quantifies the change in semantic scope, as:  
\begin{align*}
    g_{\mathrm{SUS}}(w) = \log\frac{V_T}{V_S}.
\end{align*}

\section{Experimental Setup} \label{sec:experiment_preliminary}

\subsection{Experimental configuration}
\paragraph{Dataset.}

In the experiments, we used the Diachronic Word Usage Graph (DWUG) dataset for English version\,3~\citep{schlechtweg-etal-2021-dwug, DBLP:conf/emnlp/SchlechtwegCNAW24}. It contains 46 target words. For each target word, about 100 usage instances are provided for an old time period (1810--1860) and also for a new time period (1960--2010). For details on DWUG, refer to Appendix~\ref{appx:detail-dwug}. In particular, an overview of the target word \textit{record} is provided in Table~\ref{tab:dwug-example} therein.

In addition, to verify that the proposed method does not overfit to a single dataset, we also conducted experiments on DWUG ES~\citep{DBLP:conf/acl-lchange/Zamora-ReinaBS22}, a dataset designed to detect semantic change in Spanish. This dataset was constructed following the same procedure used to construct English DWUG dataset. The results on this dataset are presented in Appendix~\ref{appx:dwug_es}.

\paragraph{Embedding extraction.}
XL-LEXEME~\citep{cassotti-etal-2023-xl} was used to obtain contextualized word embeddings. The dimension of embeddings is $d = 1024$. See Appendix~\ref{appx:detail-xl-lexeme} for details on the process of XL-LEXEME calculating the contextualized embeddings using the DWUG dataset.

\paragraph{UOT parameters.}
We set uniform weights for each usage instance:~$\bm a = \left(\frac{1}{m}, \ldots, \frac{1}{m}\right)^\top, \quad \bm b = \left(\frac{1}{n}, \ldots,\frac{1}{n}\right)^\top$. We defined the transportation cost between instances using the cosine distance:~$C_{ij} = 1 - \cos (\bm u_i, \bm v_j)$. For penalizing excess or deficit in the alignment, we used the L2 error:~$ D_1(\bm T\bm 1_n, \bm a) = \|\bm T\bm 1_n - \bm a\|_2^2$, $D_2(\bm T^\top\bm 1_m, \bm b) = \|\bm T^\top\bm 1_m - \bm b\|_2^2$. The majorization-minimization (MM) algorithm provided by Python Optimal Transport~\citep{flamary2021pot} was used for implementation. Following previous work on UOT~\citep{DBLP:conf/nips/ChapelFWFG21}, we set $\lambda_1 = \lambda_2 = \lambda$. The value of $\lambda$ was kept consistent across all target words. For qualitative analysis in Section~\ref{sec:experiment_visualization}, $\lambda$ was fixed at $\lambda = 100$. For quantitative evaluation using SUS in Sections~\ref{sec:experiment_instance-level} and \ref{sec:experiment_word-level}, $\lambda$ was determined for each experiment by using a validation set. 

Regarding the weights $\bm{a}$ and $\bm{b}$, one alternative is to use the norms of word embeddings, as proposed by \citet{DBLP:conf/emnlp/YokoiTASI20}. However, this choice had little impact on performance in the task described in Section~\ref{sec:experiment_word-level}. This is likely because all embeddings represent the same target word, and their norms do not vary substantially.

As a penalty function, the Kullback-Leibler divergence has also been used in UOT~\citep{DBLP:conf/nips/ChapelFWFG21, DBLP:conf/acl/Arase0Y23}, but we consider that using the L2 error is more natural for focusing on the excess and deficit of alignment.

\subsection{Sense frequency distribution (SFD)} \label{subsec:sfd}
\paragraph{Gold SFD.}
Here, we describe the gold standard\footnote{The asterisk ($*$) in mathematical notation used in the experiments indicates the gold standard.} used to evaluate the performance of the proposed methods and the baselines in the experiments. In the DWUG dataset, based on human annotations, each usage instance of a target word is computationally assigned a sense $k = 1, \ldots, K^*$. Note that the senses themselves are not human-annotated, as described in Appendix~\ref{appx:detail-dwug}. In this study, we regard these sense labels in the DWUG dataset as the gold senses. Since the sense inventory is the same in the old and modern corpora, we can define the sense frequency distribution (SFD), representing the frequency of each sense $k$ in both corpora. Let $X_k^*$ and $Y_k^*$ denote the total number of instances associated with sense $k$ in the old and modern corpora, respectively. The SFDs are then expressed as $X^* = (X_1^*, \ldots, X_{K^*}^*)$ and $Y^* = (Y_1^*, \ldots, Y_{K^*}^*)$. Although the SFDs derived from DWUG may not always reflect the true SFDs, we assume them to be sufficiently reliable for our analysis.

\paragraph{Baseline SFD.}
We used WiDiD~\citep{DBLP:conf/acl-lchange/PeritiFMR22}, a sense-based method, as the baseline for evaluation experiments in Sections~\ref{sec:experiment_instance-level} and \ref{sec:experiment_word-level}. \citet{periti-tahmasebi-2024-systematic} showed that among sense-based approaches, this method achieves SOTA performance in the task described in Section~\ref{sec:word-shift-absolute}. WiDiD clusters the embeddings of the target word $w$ from the old and modern corpora, $\{\bm{u}_i\}_{i=1}^m \cup \{\bm{v}_j\}_{j=1}^n$, deriving the SFDs $\hat{X} = (\hat{X}_1, \ldots, \hat{X}_{\hat{K}})$ and $\hat{Y} = (\hat{Y}_1, \ldots, \hat{Y}_{\hat{K}})$ in each corpus.

\section{Evaluation of Quantifying Instance-Level Semantic Change}
\label{sec:experiment_instance-level}

\begin{table}[!t]
\small
    \centering
    \begin{tabular}{cccc}
        \toprule
        Method & Approach& Instance-level & Sense-level\\
        \midrule
        $\tau_{\mathrm{SUS}}$ &SUS-based& \textbf{0.46} &0.83 \\
         $\tau_{\mathrm{LDR}}$ & form-based & 0.40 & 0.70\\
         $\tau_{\mathrm{WiDiD}}$ &sense-based& 0.31 & \textbf{0.84}\\
        \bottomrule
    \end{tabular}
    \caption{Performance of methods for predicting the instance-level and sense-level change score.}
    \label{tab:usage-graded-change}
\end{table}

\begin{figure}[t]
    \centering 
    \includegraphics[width=1.0\linewidth]{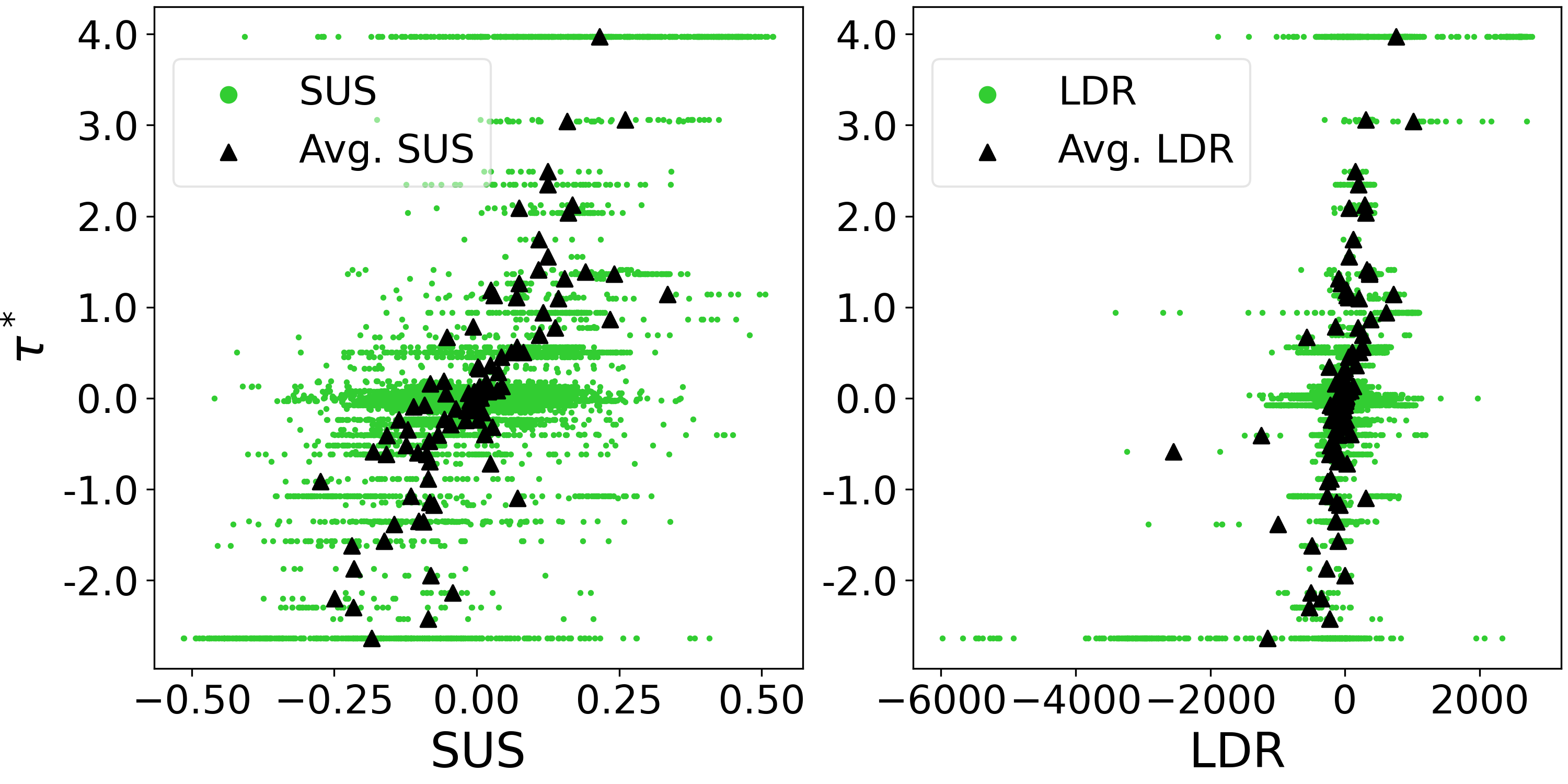}
    \caption{The relationship between the gold change score $\tau^*$ derived from DWUG and the values of SUS (left) and LDR (right) for \emph{all} instances. The average values for each $\tau^*$ are marked with $\blacktriangle$. The Spearman rank correlation computed over {\color{green} $ \bullet$} corresponds to instance-level performance and $\blacktriangle$ to sense-level performance.
    }
    \label{fig:usage-graded-plot}
\end{figure}

In this section, we investigate whether SUS is effective for the task of detecting semantic change at the individual usage instance level through numerical experiments. By using SUS, we identify the degree of change in the frequency of the word usage associated with each sense. For details of all experiments, refer to Appendix~\ref{appx:experiment_instance-level}.

The gold change score for performance evaluation and the baseline change score for comparison are constructed using a sense-based approach with SFD. While the sense-based approach can only measure semantic change at the sense level, rather than at the individual usage instance level, it serves as a proxy in the experiments.

Given two SFDs $X$ and $Y$, we define the change score\footnote{In cases where $X_k = 0$ or $Y_k = 0 $, the values are imputed using the minimum or maximum change scores across all words, respectively.} for the instance $s_i$, or $t_j$, as $\tau(s_i; X, Y) =\log \frac{Y_k}{X_k}$ where $k$ is the sense identified for $s_i$. The gold change score for a usage instance of a target word is defined as $\tau^*(\cdot) = \tau(\cdot; X^*, Y^*)$. As a baseline, using the SFDs $\hat{X}$ and $\hat{Y}$ estimated by the sense-based approach WiDiD, we calculated an estimate of $\tau^*$ as $\tau_{\mathrm{WiDiD}}(\cdot) = \tau(\cdot; \hat X, \hat Y)$. 

To estimate $\tau^*$, our proposed method directly uses the SUS value as $\tau_{\mathrm{SUS}}(\cdot) = \mathrm{SUS}(\cdot)$. As a direct baseline for SUS, we also use the LDR value as $\tau_{\mathrm{LDR}}(\cdot) = \mathrm{LDR}(\cdot)$.

Table~\ref{tab:usage-graded-change} shows the Spearman rank correlation between the gold change scores $\tau^*$ and the change score produced by each method. To clarify the situation, see Fig.~\ref{fig:usage-graded-plot}, which shows the values of SUS and LDR for all usage instances of the target words on the horizontal axis and their corresponding gold change scores $\tau^*$ on the vertical axis.
SUS outperforms the other methods, providing more accurate predictions of changes in word usage frequency for the sense of each instance.
Since instances with the same sense share the same $\tau^*$, we compute the Spearman rank correlation between the gold change scores and the average SUS values for instances with the same $\tau^*$. We call this sense-level semantic change.
SUS achieves a Spearman rank correlation of 0.83, indicating its effectiveness in capturing changes in usage frequency. In contrast, LDR shows a lower correlation of 0.70, demonstrating its inferior performance.
While WiDID performs well for sense-level semantic change detection, it is worth noting that the gold scores are based on sense-level semantic change, which inherently favors WiDID as a metric.

\section{Evaluation of Quantifying Word-Level Semantic Change}
\label{sec:experiment_word-level}

We evaluate the validity of the SUS-based metrics $f_\mathrm{SUS}$ and $g_\mathrm{SUS}$ for quantifying semantic change at the word level, as defined in Section~\ref{sec:sus-based-word-level}. Specifically, we examine the correlation between these metrics and the gold change scores derived from the gold SFDs in DWUG. 

In this section, we normalize the SFDs $X$ and $Y$ of $w$ from the old and modern corpora and denote them as $P$ and $Q$, respectively.
For two probability distributions $P$ and $Q$, we define the metrics $f$ and $g$ to quantify  unsigned and signed change, respectively, as follows:
\begin{align*}
    f(P, Q) &= \mathrm{JSD}(P, Q), \\
    g(P, Q) &= H(Q) - H(P).
\end{align*}  
Here, $\mathrm{JSD}(P, Q)$ denotes the Jensen-Shannon divergence between $P$ and $Q$, while $H$ denotes the entropy, which represents the spread of the distribution; thus, the entropy difference quantifies the change in semantic scope.

\subsection{Quantifying the magnitude of change}
\label{sec:word-shift-absolute}
\begin{table}[t]
\small
    \centering
    \begin{tabular}{ccc}
        \toprule
         Method & Approach& Spearman\\
        \midrule
         $f_{\mathrm{SUS}}$ & SUS-based & 0.69\\
        $f_\mathrm{OT}$ & form-based & \textbf{0.71}\\
        $f_{\mathrm{APD}}$ &form-based & \textbf{0.71}\\
          $f_\mathrm{LDR}$&form-based  & 0.31\\
         $f_{\mathrm{WiDiD}}$ & sense-based  & 0.45\\
         $f_{\mathrm{APDP}}$ & sense-based  & 0.51\\
        \bottomrule
    \end{tabular}
    \caption{Performance of methods for measuring the magnitude of word-level semantic change. For the complete results, refer to Table~\ref{tab:appx-JSD-rank-corr} in Appendix~\ref{appx:word-shift-absolute}.}
    \label{tab:JSD-rank-corr}
\end{table}

Following \citet{schlechtweg-etal-2020-semeval}, the gold change score is defined as $f^*(w) = f(P^*, Q^*)$ computed from the gold SFDs $X^*$ and $Y^*$.

For comparison, we used five baselines in this evaluation. As a baseline change score for the form-based approach, following \citet{periti-tahmasebi-2024-systematic}, we used the average pairwise distance (APD) denoted as $f_{\mathrm{APD}}(w)$. As a baseline change score for the sense-based approach, we used $f_{\mathrm{WiDiD}}(w)=f(\hat P, \hat Q)$ computed from the estimated SFDs $\hat X$ and $\hat Y$ in WiDiD, and also the APD between sense prototypes (APDP) yielded by clustering in WiDiD, denoted as $f_{\mathrm{APDP}}(w)$. Furthermore, we used the standard OT distance denoted as $f_\mathrm{OT}(w)$. We also defined a metric $f_\mathrm{LDR}(w)$ based on LDR of each usage instance. See Appendix~\ref{appx:word-shift-absolute} for the details of the baseline metrics.

Table~\ref{tab:JSD-rank-corr} shows the Spearman rank correlation coefficients between the gold change score $f^*$ and the change score produced by each method. The metric $f_\mathrm{APD}$, which directly utilizes embeddings, achieves the highest performance. However, the change score based on SUS demonstrates comparable performance, confirming that SUS effectively captures word-level semantic change. Furthermore, OT also achieves the highest performance, demonstrating that both OT and UOT are highly effective for detecting word-level semantic change.

\subsection{Quantifying the change in semantic scope}
\label{sec:word-shift-signed}

\begin{table}[t]
\small
    \centering
    \begin{tabular}{ccc}
        \toprule
          Method &Approach& Spearman\\
        \midrule
         $g_{\mathrm{SUS}}$ & SUS-based & 0.55\\
         $g_{\mathrm{vMF}}$ &form-based & \textbf{0.62}\\
         $g_\mathrm{LDR}$ &form-based&0.36\\
         $g_{\mathrm{WiDiD}}$ & sense-based &0.40 \\
        \bottomrule
    \end{tabular}
    \caption{Performance of methods for measuring word-level changes in semantic scope. For the complete results, refer to Table~\ref{tab:appx-ENT-rank-corr} in Appendix~\ref{appx:word-shift-signed}.}
    \label{tab:ENT-rank-corr}
\end{table}

Following \cite{DBLP:conf/acl/GiulianelliTF20}, the gold score for change in semantic scope is the entropy difference between $P^*$ and $Q^*$, i.e., $g^*(w) = g(P^*, Q^*)$.

As a baseline score for the form-based approach, following \citet{nagata-etal-2023-variance}, we used a metric called coverage denoted as $g_\mathrm{vMF}$. Moreover, we defined $g_\mathrm{LDR}$ based on LDR of each usage instance. As a sense-based approach, we used $g_\mathrm{WiDiD} = g(\hat P, \hat Q)$ computed from the estimated SFDs in WiDiD. See Appendix~\ref{appx:word-shift-signed} for the details of the baseline metrics.

Table~\ref{tab:ENT-rank-corr} shows the Spearman rank correlation coefficients between the gold change score $g^*$ and the change score calculated by each method. The form-based approach achieves the highest performance, effectively capturing the broadening or narrowing of meaning. The change score based on SUS outperforms the sense-based method, demonstrating a certain level of validity.

\section{Conclusion}
\label{sec:conclusion}

We applied UOT to sets of contextualized embeddings of a target word in a diachronic corpus pair. By leveraging the excess and deficit in the alignment between usage instances, we proposed a new measure called SUS for each usage instance to quantify changes in the frequency of word usage associated with its sense. The effectiveness of SUS was evaluated through experiments on semantic change detection tasks.

\section*{Limitations}
\begin{itemize}
    \item While SUS can detect detailed changes in individual usage instances, form-based methods that directly utilize contextualized word embeddings slightly outperform the SUS-based method for quantifying word-level semantic change.

    \item The SUS value depends on the hyperparameter $\lambda$ used in UOT. Since the optimal $\lambda$ value may vary depending on the data, further investigations are needed to establish effective $\lambda$ tuning strategies.

    \item In this paper, we used only the DWUG dataset for English. Experiments on other languages remain a topic for future work.

    \item Since UOT is applied under the constraint of handling transport between two probability distributions, changes in the total occurrence counts of a target word across a diachronic corpus pair are not considered. In other words, UOT focuses on the relative frequency of word usage associated with a sense within each corpus. Additionally, the dataset used in this study does not provide effective information about the total occurrence counts of the target words. These limitations could potentially be addressed in the future through extensions of UOT and the acquisition of datasets that include information on total occurrence counts.

\end{itemize}

\section*{Ethics Statement}
This study complies with the \href{https://www.aclweb.org/portal/content/acl-code-ethics}{ACL Ethics Policy}.

\section*{Acknowledgments}
We would like to thank the anonymous reviewers for their helpful comments and suggestions.
This study was partially supported by JSPS KAKENHI 22H05106, 23H03355, 23K24910, JST FOREST JPMJFR2331, and JST CREST JPMJCR21N3.

\bibliography{custom}

\appendix

\section{Further Illustrative Examples}
\label{appx:experiment_visualization}

\begin{figure*}[p]
  \begin{minipage}[b]{0.45\linewidth}
    \centering
    \includegraphics[width=1.0\linewidth]{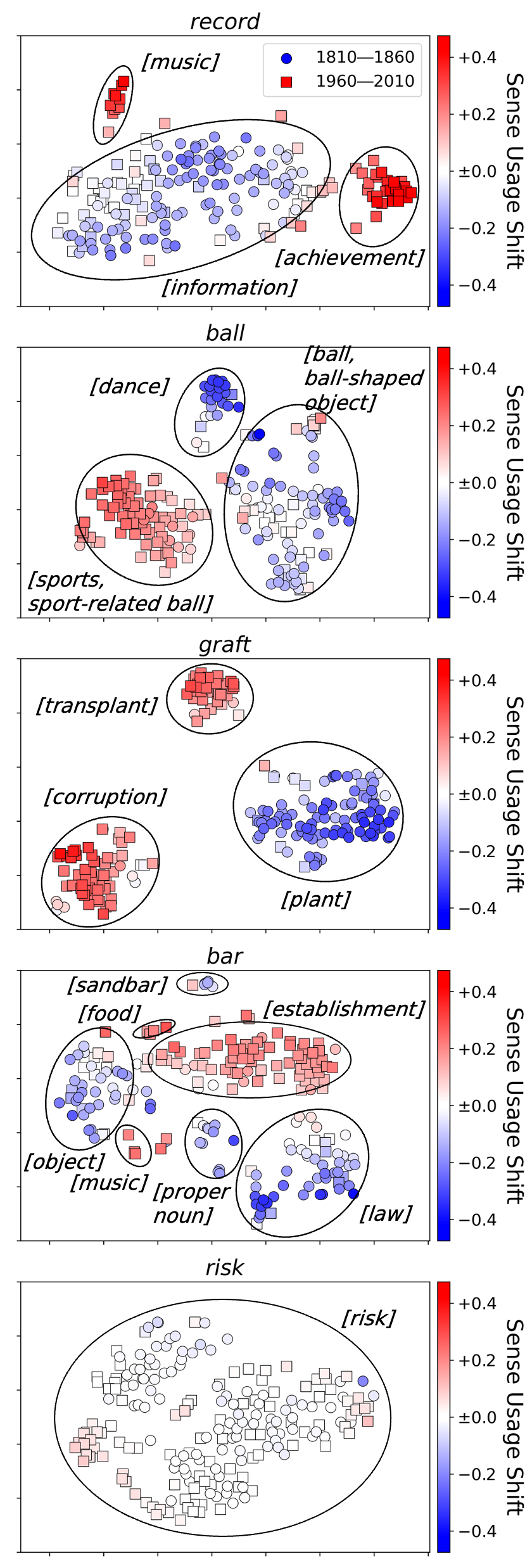}
    \subcaption{SUS}
    \label{fig:appx-sus-all}
  \end{minipage}
  \begin{minipage}[b]{0.45\linewidth}
    \centering
    \includegraphics[width=1.0\linewidth]{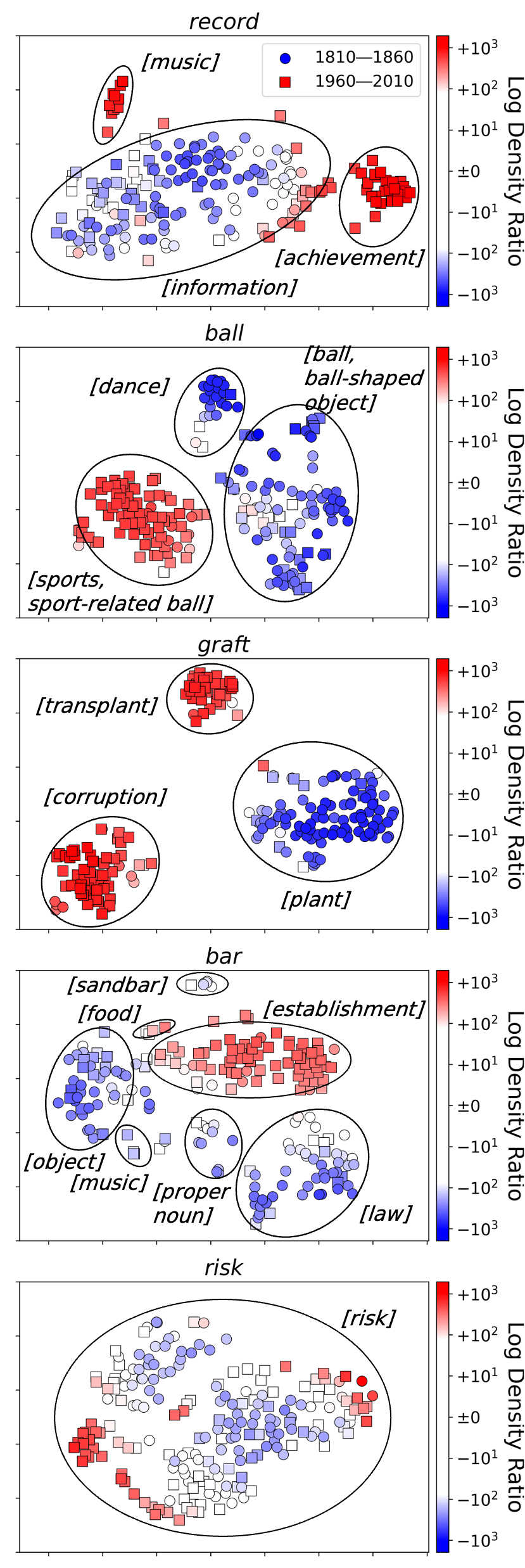}
    \subcaption{LDR}
    \label{fig:appx-ldr-all}
  \end{minipage}
  \caption{t-SNE visualization of the contextualized embeddings for each usage instance of a target word in the diachronic corpus pair. The color of each instance represents its (a) SUS or (b) LDR.}
  \label{fig:appx-sus-ldr}
\end{figure*}

\begin{table*}[p]
    \centering
    \begin{subtable}[t]{\linewidth}
        \scriptsize
        \centering
        \begin{tabular}{ccccc}
        \toprule
                                  & Corpus & Instance                                            & Sense       & SUS   \\
        \midrule
        \multirow{5}{*}{Top 5}    & 1960--2010      & So did Sire \textbf{\textit{Records}}...                              & \textit{[music]}       & 0.47  \\
                                  & 1960--2010      & ... a team with the third-worst \textbf{\textit{record}}...           & \textit{[achievement]} & 0.45  \\
                                  & 1960--2010      & ... the AMCU single-season \textbf{\textit{record}}...                & \textit{[achievement]} & 0.45  \\
                                  & 1960--2010      & ... the indoor \textbf{\textit{record}} of 24-2 1/2...                & \textit{[achievement]} & 0.44  \\
                                  & 1960--2010      & ... the late Steve Prefontaine's American \textbf{\textit{record}}... & \textit{[achievement]} & 0.44  \\
        \midrule
        \multirow{5}{*}{Bottom 5} & 1810--1860      & This \textbf{\textit{record}} shall be read at the commencement...    & \textit{[information]} & -0.22 \\
                                  & 1810--1860      & ... in the comprehensive \textbf{\textit{records}} of philosophy...   & \textit{[information]} & -0.23 \\
                                  & 1810--1860      & ... interpretations of the Mosaic \textbf{\textit{record}}...         & \textit{[information]} & -0.24 \\
                                  & 1810--1860      & ... the \textbf{\textit{records}} of a professed revelation...        & \textit{[information]} & -0.24 \\
                                  & 1810--1860      & ... the \textbf{\textit{record}} of whose wisdom is included in...    & \textit{[information]} & -0.25 \\
        \bottomrule
        \end{tabular}
        \caption{\textit{record}}
        \label{tab:appx-record-sus-example}
        \vspace{0.3cm}
    \end{subtable}

    \begin{subtable}[t]{\linewidth}
        \scriptsize
        \centering
        \begin{tabular}{ccccc}
        \toprule
                                  & Corpus     & Instance                                      & Sense                       & SUS   \\ 
        \midrule
        \multirow{5}{*}{Top 5}    & 1960--2010 & ... by teaching Wagner a palm \textbf{\textit{ball}}.           & \textit{[sport-related ball]} & 0.31  \\
                                  & 1960--2010 & ... flip the \textbf{\textit{ball}} to a trailing halfback...   & \textit{[sport-related ball]} & 0.27  \\
                                  & 1960--2010 & ... a safety pass, and if he gets the \textbf{\textit{ball}}... & \textit{[sport-related ball]} & 0.27  \\
                                  & 1960--2010 & ... pass the \textbf{\textit{ball}} to a spot...                & \textit{[sport-related ball]} & 0.26  \\
                                  & 1960--2010 & ... you run that \textbf{\textit{ball}} again you're out...     & \textit{[sport-related ball]} & 0.26  \\
        \midrule
        \multirow{5}{*}{Bottom 5} & 1810--1860 & ... at a tea-party, or a \textbf{\textit{ball}}...              & \textit{[dance]}              & -0.32 \\
                                  & 1810--1860 & ... at, a \textbf{\textit{ball}}, or dance...                   & \textit{[dance]}              & -0.32 \\
                                  & 1810--1860 & I now began to attend \textbf{\textit{balls}}...                & \textit{[dance]}              & -0.34 \\
                                  & 1810--1860 & It is a masked \textbf{\textit{ball}}...                        & \textit{[dance]}              & -0.34 \\
                                  & 1810--1860 & ... keep up the \textbf{\textit{ball}} of conversation...       & \textit{[ball, ball-shaped]}  & -0.42 \\
        \bottomrule
        \end{tabular}
        \caption{\textit{ball}}
        \label{tab:appx-ball-sus-example}
        \vspace{0.3cm}
    \end{subtable}
    
    \begin{subtable}[t]{\linewidth}
        \scriptsize
        \centering
        \begin{tabular}{ccccc}
        \toprule
              & Corpus                      & Instance                                                     & Sense      & SUS   \\
        \midrule
        \multirow{5}{*}{Top 5}    & 1960--2010                  & ... political wheeling and dealing and \textbf{\textit{graft}}...              & \textit{[corruption]} & 0.40  \\
                                  & 1960--2010                  & ... sweetheart contracts, and outright \textbf{\textit{graft}}.                & \textit{[corruption]} & 0.38  \\
                                  & 1960--2010                  & ... there is less \textbf{\textit{graft}} in the police department...          & \textit{[corruption]} & 0.36  \\
                                  & 1960--2010                  & ... an average person would consider to be \textbf{\textit{graft}}.            & \textit{[corruption]} & 0.34  \\
                                  & 1960--2010                  & ... political \textbf{\textit{graft}} is rampant in every department...        & \textit{[corruption]} & 0.34  \\
        \midrule
        \multirow{5}{*}{Bottom 5} & 1810--1860 & ... between the stock and \textbf{\textit{graft}}... from a lofty tree...      & \textit{[plant]}      & -0.33 \\
                                  & 1810--1860 & In the following spring, the \textbf{\textit{grafted}} trees...                & \textit{[plant]}      & -0.33 \\
                                  & 1810--1860 & ... \textbf{\textit{grafts}} of a few varieties inserted at standard height... & \textit{[plant]}      & -0.33 \\
                                  & 1810--1860 & ... which have been \textbf{\textit{grafted}} late in spring...                & \textit{[plant]}      & -0.35 \\
                                  & 1810--1860 & ... a tree but one year from the \textbf{\textit{graft}}...                    & \textit{[plant]}      & -0.35\\
        \bottomrule
        \end{tabular}
        \caption{\textit{graft}}
        \label{tab:appx-graft-sus-example}
        \vspace{0.3cm}
    \end{subtable}

    \begin{subtable}[t]{\linewidth}
        \scriptsize
        \centering
        \begin{tabular}{ccccc}
        \toprule
                                  & Corpus     & Instance                                                       & Sense                  & SUS   \\ \midrule
        \multirow{5}{*}{Top 5}    & 1960--2010 & ... from the snack \textbf{\textit{bar}}...                                      & \textit{[establishment]} & 0.29  \\
                                  & 1960--2010 & He'd lined up the bottles on the \textbf{\textit{bar}}...                        & \textit{[establishment]} & 0.26  \\
                                  & 1960--2010 & ... but since the window lacks scroll \textbf{\textit{bars}}...                  & \textit{[object]}        & 0.26  \\
                                  & 1960--2010 & ... the oboe to rest for a certain number of \textbf{\textit{bars}}...           & \textit{[music]}         & 0.24  \\
                                  & 1960--2010 & Champagne, hors d'oeuvres, a five-course dinner, open \textbf{\textit{bar}}...   & \textit{[establishment]} & 0.24  \\ \midrule
        \multirow{5}{*}{Bottom 5} & 1810--1860 & ... gentlemen of the \textbf{\textit{bar}}... constant attendance upon courts... & \textit{[law]}           & -0.33 \\
                                  & 1810--1860 & ... some member of the \textbf{\textit{bar}}...                                  & \textit{[law]}           & -0.34 \\
                                  & 1810--1860 & Before a legal tribunal... at the \textbf{\textit{bar}} of public opinion...     & \textit{[law]}           & -0.35 \\
                                  & 1810--1860 & ... before the \textbf{\textit{bar}} of Judgment...                              & \textit{[law]}           & -0.35 \\
                                  & 1810--1860 & \textbf{\textit{bar}} of the House, against the matter of the charges...         & \textit{[law]}           & -0.40 \\ \bottomrule
        \end{tabular}
        \caption{\textit{bar}}
        \label{tab:appx-bar-sus-example}
        \vspace{0.3cm}
    \end{subtable}

    \begin{subtable}[t]{\linewidth}
        \scriptsize
        \centering
        \begin{tabular}{ccccc}
        \toprule
                                  & Corpus     & Instance                                                       & Sense         & SUS    \\ \midrule
        \multirow{5}{*}{Top 5}    & 1960--2010 & ... to take \textbf{\textit{risks}} with executives...                           & \textit{[risk]} & 0.11   \\
                                  & 1960--2010 & Being overweight increases your \textbf{\textit{risk}}...                        & \textit{[risk]} & 0.090  \\
                                  & 1960--2010 & ... people at high \textbf{\textit{risk}} of breast cancer...                    & \textit{[risk]} & 0.090  \\
                                  & 1960--2010 & ... an increased \textbf{\textit{risk}} of type 2 diabetes.                      & \textit{[risk]} & 0.064  \\
                                  & 1960--2010 & ... have an increased \textbf{\textit{risk}} of breast cancer.                   & \textit{[risk]} & 0.060  \\ \midrule
        \multirow{5}{*}{Bottom 5} & 1810--1860 & ... at the \textbf{\textit{risk}} of holding him too long from...                & \textit{[risk]} & -0.058 \\
                                  & 1810--1860 & ... at the \textbf{\textit{risk}} of her life...                                 & \textit{[risk]} & -0.062 \\
                                  & 1810--1860 & At the \textbf{\textit{risk}} of repeating what may be already quite familiar... & \textit{[risk]} & -0.064 \\
                                  & 1810--1860 & ... at the \textbf{\textit{risk}} of arguing ourselves unknown...                & \textit{[risk]} & -0.069 \\
                                  & 1810--1860 & ... must charge interest and \textbf{\textit{risk}}...                           & \textit{[risk]} & -0.20  \\ \bottomrule
        \end{tabular}
        \caption{\textit{risk}}
        \label{tab:appx-risk-sus-example}
    \end{subtable}

    \caption{The top 5 and bottom 5 usage instances of the target words based on SUS values.}
    \label{tab:appx-sus-tables}
\end{table*}

\begin{figure}[p]
    \centering
    \includegraphics[width=1.0\linewidth]{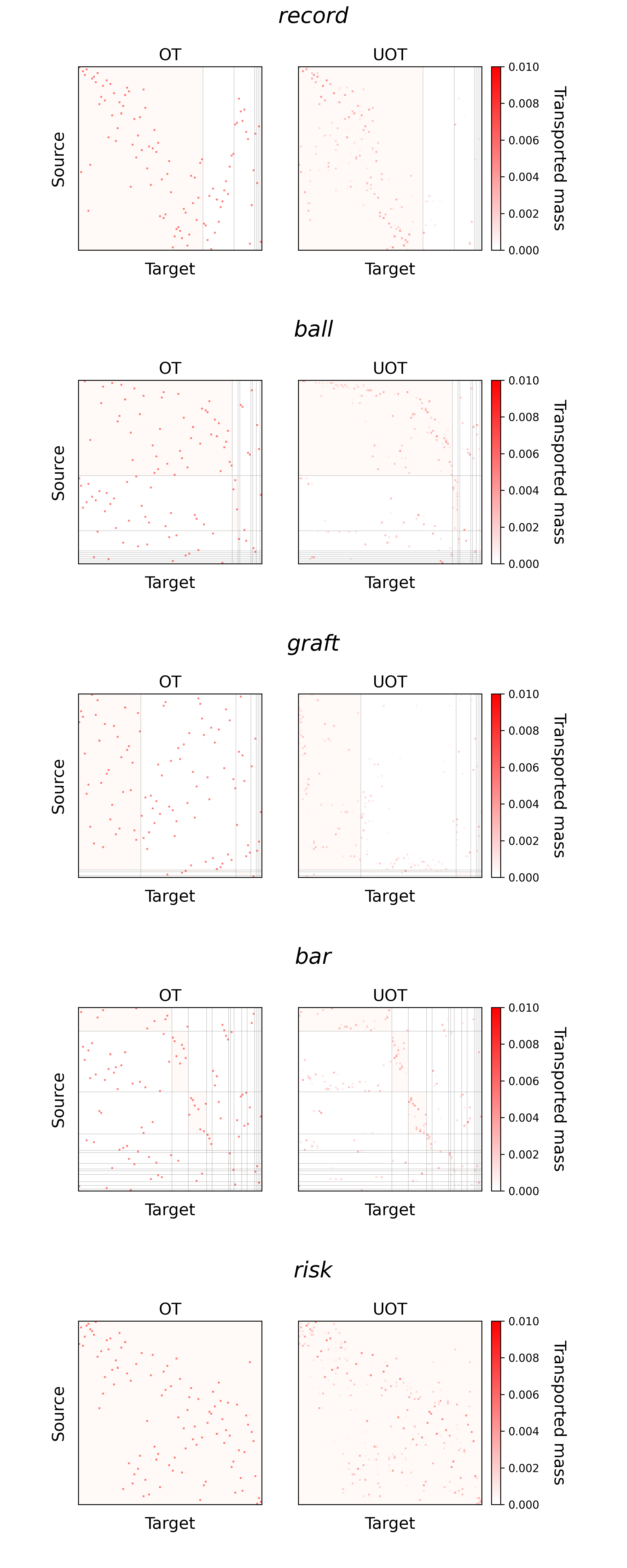}
    \caption{The transportation matrices of OT and UOT for contextualized embeddings from old and modern corpora. Red-shaded areas indicate alignments between instances in the same word sense across the two corpora, while white areas indicate alignments between different senses. By focusing on the white blocks, it is observed that OT conducts transportation across different senses, whereas UOT reduces such transportation by allowing alignment discrepancy.}
    \label{fig:appx-OT-UOT-matrix}
\end{figure}

\paragraph{Visualizing semantic change with SUS and LDR.}
We selected five target words: \textit{record}, \textit{ball}, \textit{graft}, \textit{bar}, and \textit{risk}. The semantic change visualizations for the five target words are presented in Fig.~\ref{fig:appx-sus-ldr}. Note that the target word \textit{risk} does not exhibit semantic change in the dataset. For color-coding each instance based on its values (SUS or LDR), the maximum of the 95\% percentiles and the minimum of the 30\% percentiles of the absolute values for each target word were used as thresholds. Instances with values exceeding the upper threshold are highlighted with the darkest colors (red or blue), while those below the lower threshold are rendered in white. Additionally, since LDR takes very large values in high-dimensional settings, the transformation $ \mathrm{sgn}(\mathrm{LDR}) \log(1+|\mathrm{LDR}|) $ was applied before assigning colors.

\paragraph{Comparison of SUS and LDR Visualizations.}
As described in Section~\ref{sec:ldr}, Fig.~\ref{fig:appx-sus-ldr} indicates that both SUS and LDR generally capture the sense clusters reasonably well. However, a more detailed examination highlights distinctions between the two approaches. For the target word \textit{record}, when compared against the gold senses from the DWUG dataset shown in Fig.~\ref{fig:appx-record-tsne-dwug}, SUS appears to align more closely with the actual clusters than LDR. Specifically, in the rightmost portion of the cluster representing the sense \textit{[information]}, there is a region where circles and squares overlap. LDR distinguishes this area as separate clusters more clearly than SUS does.
For the target word \textit{bar}, the sense \textit{[music]} newly emerged in the modern corpus. SUS correctly identifies this increase in usage frequency, whereas LDR incorrectly identifies it as a frequency decrease. However, such detailed observations have not been sufficiently discussed, and caution is needed when drawing conclusions from these differences.  

Another noteworthy difference lies in the interpretability of the SUS and LDR values. The SUS values range from $-1$ to $+1$, making it easier to interpret their magnitude. In contrast, LDR values fluctuate over a much larger range, often varying substantially, making it difficult to interpret their magnitude. For example, for the target word \textit{risk}, which has only one sense \textit{[risk]}, SUS correctly identifies that no semantic change has occurred. In contrast, LDR identifies frequency changes in unnecessarily small regions, dividing what should be a single cluster.

\paragraph{Instances with high or low SUS values.}
Additionally, Table~\ref{tab:appx-sus-tables} shows the usage instances of the target words with the top 5 and bottom 5 SUS values. For example, Table~\ref{tab:appx-record-sus-example} presents the instances of \textit{record} with high and low SUS values. The senses of the top 5 instances correspond to those whose usage frequencies have increased in the modern corpus, such as \textit{[music]} and \textit{[achievement]}. Conversely, Table~\ref{tab:appx-ball-sus-example} presents the instances of \textit{ball}, where the senses of the bottom 5 instances correspond to those whose usage frequencies have decreased in the modern corpus, including \textit{[dance]} and \textit{[ball, ball-shaped object]}.
As discussed in Section~\ref{sec:experiment_visualization}, the meaning of \textit{record} has broadened while the meaning of \textit{ball} has narrowed across the two corpora. Note that word usage in the word sense \textit{[dance]} remains common today. However, our analysis focuses on comparing the given two corpora which have the same number of instances sampled from a larger corpus.
Therefore, as mentioned in the Limitations section, while we can quantify changes in the relative frequency of word usage for specific senses, we cannot capture changes in the total count of word occurrences.

\paragraph{Difference between OT and UOT.}
Figure~\ref{fig:appx-OT-UOT-matrix} visualizes the transportation matrices of OT and UOT between the sets of the contextualized embeddings from the old and modern corpora for several target words. The usage instances are sorted according to the gold senses provided in DWUG. Within each sense, the instances are further ordered by their $x$-values in their t-SNE visualizations. Lines indicate the boundaries between different senses.
The $(i, j)$ element belongs to the red-shaded blocks if the senses of $s_i$ and $t_j$ are the same, and it belongs to the white blocks if the senses are different.
In the case of OT in Fig.~\ref{fig:appx-OT-UOT-matrix}, it is observed that a substantial amount of transport occurs within the white blocks. This indicates that alignments are conducted between the instances with different senses across the two corpora. As described in Section~\ref{sec:OT-limitation}, this happens because OT enforces a balanced alignment between the instances from the two corpora, which fails to fully capture semantic change at the instance level. 
On the other hand, for UOT, the transported mass within the white blocks is observed to be small, indicating that transportation across instances with different senses is avoided.
This is because, as discussed in Section~\ref{sec:UOT-proposed}, UOT allows for excess or deficit of alignment.
Thus, UOT more effectively captures semantic change at the instance level.

\section{Details of Log-Density Ratio}
\label{appx:ldr}

\paragraph{Computation.}
In our experiments, we directly computed LDR from Eq.~(8) of \citet{nagata-etal-2023-variance}. Specifically, we calculated the approximate maximum likelihood estimator (MLE) of the concentration parameter $\kappa$ for the von Mises-Fisher (vMF) distribution for the old and modern corpora as follows:
Let $\{\bm u_i\}_{i=1}^m \subset \mathbb{R}^d$ denote the set of embeddings in the old corpus and define $\tilde{\bm u}_i = \bm u_i /\|\bm u_i\|$. Assume that $\{\tilde{\bm u}_i\}_{i=1}^m$ follow a vMF distribution with mean direction parameter $\bm\mu_S$ and concentration parameter $\kappa_S$. Let $\ell = \left\|\frac{1}{m}\sum_i \tilde{\bm u}_i\right\|$. Then, the maximum likelihood estimator (MLE) of $\bm\mu_S$ is given by $\frac{1}{\ell m}\sum_i \tilde{\bm u}_i$. While the MLE of $\kappa_S$ does not have a closed-form solution, \citet{DBLP:journals/jmlr/BanerjeeDGS05} provide the following approximation: $\kappa_S \approx \frac{\ell(d - \ell^2)}{1 - \ell^2}$.  
After calculating the MLEs, we derived the corresponding vMF density function using \texttt{vonmises\_fisher} in \texttt{scipy}.  
We applied the same procedure to the modern embeddings $\{\bm v_j\}_{j=1}^n$ by computing their normalized forms $\tilde{\bm v}_j = \bm v_j / \|\bm v_j\|$ and estimating the parameters $\bm\mu_T$ and $\kappa_T$ of the associated vMF distribution.

\paragraph{Representativeness.}
In \citet{nagata-etal-2023-variance}, an approximation of LDR with the constant term omitted is introduced as \emph{representativeness} in Eq.~(3), which is proposed as a measure for extracting typical word instances. Due to the omission of the constant term, the sign of \emph{representativeness} does not indicate whether usage has increased or decreased.

\section{Detailed Settings for UOT}
\label{appx:uot_detail}

\subsection{Computation of UOT}

The Unbalanced Optimal Transport (UOT) problem is formulated as \eqref{eq:UOT}.
In general, if the objective function is convex, the Majorization-Minimization (MM) algorithm can be applied. Specifically, when $D_1$ and $D_2$ are Bregman divergences\footnote{Examples include  KL divergence, L1 error, and L2 error.}, they are convex with respect to their first arguments, making the objective function convex. This implies that the optimal solution to \eqref{eq:UOT} can be computed using the MM algorithm~\citep{DBLP:conf/nips/ChapelFWFG21}. The standard OT is recovered when $\lambda_1=\lambda_2\to\infty$.

\subsection{Determining the range for hyperparameter tuning}
\begin{table*}[!t]
\small
\centering
\begin{tabular}{ccccccc}
\toprule
& \multirow{2}{*}{Gold metric}            & \multirow{2}{*}{Proposed metric} & \multicolumn{4}{c}{$\lambda$}                             \\
\cmidrule(lr){4-7}
&                                 &                               & 1          & 10         & 100        & 1000       \\
\midrule
instance-level& $\tau^*$                & $\tau_{\mathrm{SUS}}$                           & 0.31 & 0.42 & \textbf{0.50} & 0.49\\
\midrule
\multirow{6}{*}{word-level}&\multirow{4}{*}{$f^*$ } &$f_\mathrm{SUS}$&0.54 &0.73 &\textbf{0.76} &\textbf{0.76}  \\
                               & & $f_1$                            & 0.42       & \textbf{0.73}       & 0.72       & 0.70        \\
                                &  &   $f_2$                          & 0.48 (0.9) & \textbf{0.78} (0.8) & 0.75 (0.2) & 0.74 (0.2) \\
                                & & $f_3$                            & -0.55      & -0.23      & \textbf{0.76}       & \textbf{0.76}       \\
\cmidrule(lr){2-7}
& \multirow{2}{*}{$g^*$}                 &$g_\mathrm{SUS}$& -0.06 & 0.57 & \textbf{0.59} & 0.56\\
 &                                                 & $g_1$                         & 0.00 (0.9)    & 0.54 (0.8) & \textbf{0.57} (0.5) & 0.55 (0.4) \\
\bottomrule
\end{tabular}
\caption{Performance of each task using all target words. The performance in the tasks using gold metrics $f^*$, $g^*$, and $\tau^*$ is reported as Spearman rank correlations.
For methods involving the threshold $\theta$, the optimal value of $r$ is provided in parentheses alongside the performance.}
\label{tab:all-words}
\end{table*}

The hyperparameter $\lambda_1 = \lambda_2 = \lambda$ in UOT \eqref{eq:UOT} needs to be tuned on a validation set during the performance evaluations conducted in Sections~\ref{sec:experiment_instance-level} and \ref{sec:experiment_word-level}. To determine the range of $\lambda$ values to explore, we initially evaluated the performance using all target words without splitting the validation and test sets. The results of varying $\lambda$ across the values 1, 10, 100, and 1000 are presented in Table~\ref{tab:all-words}. Note that additional metrics $f_1, f_2, f_3$ and $g_1$ for quantifying the degree of semantic change have been proposed as defined in Appendix~\ref{appx:experiment_word-level}.

Additionally, since the range of SUS values for each target word varies depending on $\lambda$, the threshold $\theta$ used in $f_2$ and $g_1$ is determined as a proportion of the maximum absolute SUS value within a given set of target words (valid, test, or all). Specifically, let $M$ denote the maximum value of $|\mathrm{SUS}(s_i)|$ or $|\mathrm{SUS}(t_j)|$ for $i = 1, \ldots, m$, $j = 1, \ldots, n$, and $w \in W$, where $W$ is the set of target words. The threshold is then defined as $\theta = M r$, where $r$ is the proportion to explore. In other words, instead of directly tuning $\theta$, we search for $r$. Initially, $r$ is varied across values $0.1, 0.2, \ldots, 0.9$ to establish a baseline range. Table~\ref{tab:all-words} presents the optimal $r$ for each value of $\lambda$.

Based on these results, when the only hyperparameter to be tuned was $\lambda$, we explored $\lambda \in \{10, 20, 50, 100, 200, 500, 1000\}$. When both $\lambda$ and $r$ were tuned, we varied $\lambda \in \{10, 100, 1000\}$ and $r \in \{0.4, 0.6, 0.8\}$.

\subsection{Details of hyperparameter tuning in Sections~\ref{sec:experiment_instance-level} and \ref{sec:experiment_word-level}}
\begin{table}[t]
\small
    \centering
    \begin{tabular}{ccc}
    \toprule
    Proposed metric     & Selected $\lambda$ or $(\lambda, r)$ & Times \\
    \midrule
    $\tau_{\mathrm{SUS}}$ & 100 & 73 \\
     \midrule
    $f_\mathrm{SUS}$ & 20 & 51\\
     $f_1$ & 20 & 90 \\ 
    $f_2$ & (10, 0.8) & 78 \\
    $f_3$ & 100 & 46 \\
    \midrule
    $g_\mathrm{SUS}$ & 50 & 54\\
     $g_1$ & (10, 0.8) & 53\\
    \bottomrule
    \end{tabular}
    \caption{The most frequently selected hyperparameters for the proposed instance-level and word-level metrics in the validation set, where train-validation splits are conducted 100 times.}
    \label{tab:selected-hyperparameter}
\end{table}

The hyperparameter $\lambda$ is tuned on the validation set for each performance evaluation task. The detailed procedure is outlined as follows:

\paragraph{Tuning of $\lambda$.}
The 46 target words in the DWUG dataset are randomly split into validation and test sets in an 8:2 ratio. For the validation set, the optimal $\lambda$ is determined by exploring $\lambda \in \{10, 20, 50, 100, 200, 500, 1000\}$. In the case of WiDiD, a hyperparameter called damping (ranging in $[0.5, 1.0)$) is tuned by exploring $\{0.5, 0.6, 0.7, 0.8, 0.9\}$. Subsequently, performance evaluation is conducted on the test set. This validation-test splitting is repeated 100 times, and the final performance of each method is reported as the average test performance across all splits.

\paragraph{Tuning of $\lambda$ and $r$.}
The 46 target words in the DWUG dataset are randomly split into validation and test sets in a ratio of 8:2. For the validation set, the optimal $\lambda$ and $r$ are determined by exploring $\lambda \in \{10, 100, 1000\}$ and $r \in \{0.4, 0.6, 0.8\}$. The subsequent procedure follows the same steps as described above.

\paragraph{Selected hyperparameters.}
Table~\ref{tab:selected-hyperparameter} shows the most frequently selected hyperparameter values for each evaluation task.

\section{Details of DWUG}
\label{appx:detail-dwug}

\begin{table*}[t]
\small
\centering
\begin{tabular}{cccc}
\toprule
Corpus & Instance & Sense  & Gold SFD \\
\midrule
\multirow{3}{*}{1810--1860} &... in all the \textbf{\textit{records}} of sorrow...  & \textit{[information]} &  \multirow{3}{*}{$X^* = (99, 0, 0, 0, 0, 0, 0)$}\\
 &... in the \textbf{\textit{record}} of any age or country... & \textit{[information]} & \\
 &... the historic \textbf{\textit{records}} of Christianity...& \textit{[information]} & \\

\midrule
\multirow{3}{*}{1960--2010} & ... this is the \textbf{\textit{record}} of my life... & \textit{[information]} & \multirow{3}{*}{$Y^*=(64, 17, 11, 1, 1, 1, 1)$} \\
  & ... single-season \textbf{\textit{record}} held by... & \textit{[achievement]}  & \\
 & The \textbf{\textit{record}} labels' new service... & \textit{[music]}\\
\bottomrule
\end{tabular}
\caption{Three usage instances each from the old and modern corpora for the target word \textit{record} included in the DWUG dataset. In DWUG, the gold Sense Frequency Distribution (SFD) is given. It represents the frequency of the target word usage with each sense. Upon reviewing the senses corresponding to each index, we identified them as \textit{[information]}, \textit{[achievement]}, \textit{[music]}, and four noise senses.}
\label{tab:dwug-example}
\end{table*}

In this appendix, we provide additional details on how DWUG~\citep{schlechtweg-etal-2021-dwug, DBLP:conf/emnlp/SchlechtwegCNAW24} constructs the gold senses for a target word, which were briefly introduced in Section~\ref{sec:experiment_preliminary}. In the DWUG dataset\footnote{The dataset is licensed under the CC BY-ND 4.0. \url{https://zenodo.org/records/14028531}}, a sufficient number of usage instance pairs for each target word are annotated by human annotators with a four-level similarity score, called the DURel relatedness scale~\citep{DBLP:conf/naacl/SchlechtwegWE18}. DWUG clusters the instances, represented as vertices in a network, based on the similarity scores, which are used as edge weights. Through this clustering, the sense of each usage instance is identified. In other words, DWUG can be interpreted as one of a sense-based approach via human-annotated similarities, not using the similarities obtained by contextualized embeddings. Since the clustering algorithm is applied, the gold SFDs may contain noise senses as shown in Table~\ref{tab:dwug-example}.

Due to the specification of the DWUG dataset, a small number of usage instances for each target word are not assigned a sense. In this paper, we refer to such senses as \textit{undefined}. However, in cases where the sense can be easily identified by the authors of this paper, the missing senses are annotated, such as \textit{[music]} in Table~\ref{tab:record-sus-example}.

\section{Details of XL-LEXEME}
\label{appx:detail-xl-lexeme}

We describe how XL-LEXEME\footnote{\url{https://huggingface.co/pierluigic/xl-lexeme}}~\citep{cassotti-etal-2023-xl} calculates the embeddings of target words. XL-LEXEME is a model obtained by fine-tuning XLM-RoBERTa~\citep{DBLP:conf/acl/ConneauKGCWGGOZ20} on Word-in-Context~\citep{DBLP:conf/naacl/PilehvarC19} and it has 561M parameters. In the DWUG dataset, each usage instance of a target word includes positional information specifying the position of the target word within its context. XL-LEXEME takes this context and positional information as input to compute the embedding for the target word. Specifically, special tokens are inserted before and after the target word in the context, and the embedding of the target word is computed as the mean of the embeddings of all tokens in the context.

\section{Details of Experiment in Section~\ref{sec:experiment_instance-level}}
\label{appx:experiment_instance-level}
\begin{table}[!ht]
    \centering
\resizebox{\linewidth}{!}{
    \begin{tabular}{ccccc}
        \toprule
        Method & Approach& Stable (25) & Changed (21) & Overall (46)\\
        \midrule
        $\tau_{\mathrm{SUS}}$ &SUS-based&\textbf{0.21} &\textbf{0.61} &  \textbf{0.46} \\
         $\tau_{\mathrm{LDR}}$ & form-based &0.10 &0.34 & 0.40   \\
         $\tau_{\mathrm{WiDiD}}$ &sense-based&0.15  &0.41 & 0.31 \\
        \bottomrule
    \end{tabular}}
    \caption{Performance of methods for predicting the instance-level change score. Overall corresponds to `Instance-level' in Table~\ref{tab:usage-graded-change}.}
    \label{tab:appx-usage-graded-change-devide}
\end{table}

\begin{table}[!ht]
    \centering
\resizebox{\linewidth}{!}{
    \begin{tabular}{ccccc}
        \toprule
        Method & Approach& Stable (25) & Changed (21) & Overall (46)\\
        \midrule
        $\tau_{\mathrm{SUS}}$ &SUS-based&\textbf{0.70} &0.85 &  0.83 \\
         $\tau_{\mathrm{LDR}}$ & form-based &0.64 &0.38 &0.70    \\
         $\tau_{\mathrm{WiDiD}}$ &sense-based& 0.54 &\textbf{0.93} &\textbf{0.84}  \\
        \bottomrule
    \end{tabular}}
    \caption{Performance of methods for predicting the sense-level change score. Overall corresponds to `Sense-level' in Table~\ref{tab:usage-graded-change}.}
    \label{tab:appx-sense-graded-change-devide}
\end{table}

For computing SUS, the hyperparameter $\lambda$ in \eqref{eq:UOT} was tuned by splitting the 46 target words in the DWUG dataset into a validation set and a test set, using the validation set for optimization. For further details, refer to Appendix~\ref{appx:uot_detail}. The selected values of $\lambda$ are provided in Table~\ref{tab:selected-hyperparameter} therein.

\paragraph{Gold change score.}
The gold graded change score for a usage instance of a target word is defined as $\tau^*(\cdot) = \tau(\cdot; X^*, Y^*)$, where $X^*$ and $Y^*$ are the gold SFDs from DWUG.

\paragraph{Proposed change score.}
We used the SUS value for each instance as an estimator of the gold graded change score $\tau^*$. Specifically, we defined $\tau_{\mathrm{SUS}}(\cdot) = \mathrm{SUS}(\cdot)$.

\paragraph{Baseline change score.}
Using the SFDs $\hat{X}$ and $\hat{Y}$ estimated by the sense-based approach WiDiD, we calculated an estimate of $\tau^*$ as $\tau_{\mathrm{WiDiD}}(\cdot) = \tau(\cdot; \hat X, \hat Y)$. Moreover, we used LDR for each usage instance directly, denoted as $\tau_{\mathrm{LDR}}(\cdot) = \mathrm{LDR}(\cdot)$.

\paragraph{Results.}
In the DWUG dataset, each target word is annotated with a label indicating whether its meaning has changed, based on old and modern SFDs. We divide the test set words according to this change annotation and report the instance-level and sense-level performances for each group in Table~\ref{tab:appx-usage-graded-change-devide} and Table~\ref{tab:appx-sense-graded-change-devide}, respectively. The overall score is computed without splitting the test set and corresponds to Table~\ref{tab:usage-graded-change}.
Hyperparameter tuning was performed using the validation set without splitting words into stable or changed categories. Performance evaluation was then conducted separately for stable and changed words in the test set.

For stable words, instance-level evaluation performs quite poorly across all methods. In stable words, semantic change is minimal, but the values of each metric still fluctuate, and even small variations in these values can result in significant changes in rankings. Therefore, instance-level rank correlation is not the most appropriate evaluation metric in this context. In any way, SUS demonstrates superior instance- and sense-level performance compared to LDR not only overall but also within both the stable and changed word categories.

\section{Details of Experiment in Section~\ref{sec:experiment_word-level}}
\label{appx:experiment_word-level}

\subsection{Quantifying the magnitude of word-level semantic change}
\label{appx:word-shift-absolute}

\begin{table}[t]
\small
    \centering
    \begin{tabular}{ccc}
        \toprule
         Method & Approach& Spearman\\
        \midrule
         $f_{\mathrm{SUS}}$ & SUS-based & 0.69\\
        $f_1$  & SUS-based& 0.68  \\
        $f_2$ & SUS-based &  0.68\\
        $f_3$ & SUS-based &  0.69\\
        $f_{\mathrm{OT}}$ &form-based & \textbf{0.71}\\
        $f_{\mathrm{APD}}$ &form-based & \textbf{0.71}\\
        $f_\mathrm{LDR}$&form-based  & 0.31\\
         $f_{\mathrm{WiDiD}}$ & sense-based  & 0.45\\
        $f_{\mathrm{APDP}}$ & sense-based  & 0.51\\
        \bottomrule
    \end{tabular}
    \caption{Performance of methods for measuring the magnitude of word-level semantic change.}
    \label{tab:appx-JSD-rank-corr}
\end{table}

\paragraph{Gold change score.}
Following \citet{schlechtweg-etal-2020-semeval}, the gold change score is calculated by the metric $f$ defined as $f^*(w) = f(P^*, Q^*)$.

\paragraph{Proposed change scores.}
Although in Section~\ref{sec:sus-based-word-level}, we only presented $f_\mathrm{SUS}$ for brevity, we also designed other metrics:
\begin{align*}
     f_{\mathrm{SUS}}(w) &=  \left| \frac{1}{m}\sum_{i=1}^m \alpha_i- \frac{1}{n}\sum_{j=1}^n \beta_j\right|,\\
    f_1(w) &= \sum_i |\alpha_i| +\sum_j |\beta_j|,\\
    f_2 (w;\theta) &= -\sum_{i: \alpha_i<-\theta} \alpha_i + \sum_{j: \beta_j>\theta} \beta_j,\\
    f_3(w) &= \sum_{i,j} C_{ij}T_{ij}. 
\end{align*}
Here, we defined $\alpha_i = \mathrm{SUS}(s_i)$ and $\beta_j = \mathrm{SUS}(t_j)$. The second measure, $f_1$, quantifies the magnitude of semantic change by summing the micro-level changes across all instances, as represented by the SUS values. The third measure, $f_2(\cdot; \theta)$, focuses on instances with large SUS values, emphasizing significant changes by introducing a threshold $\theta$ to filter out smaller SUS values. The fourth measure, $f_3$, corresponds to the transportation distance, which is the total cost of aligning the embeddings. Note that $f_{\mathrm{APD}}$ is derived from $f_3$ by setting $T_{ij} = 1/mn$.
The threshold $\theta$ is a hyperparameter. For details on the tuning process for $\lambda$ and $\theta$, see Appendix~\ref{appx:uot_detail}.

\paragraph{Baseline change scores.}
As a form-based approach, following \citet{periti-tahmasebi-2024-systematic}, we used the average pairwise distance (APD) based on cosine distance between the embeddings $\{\bm{u}_i\}_i$ and $\{\bm{v}_j\}_j$\footnote{Here, if we normalize the embeddings as $\bm{\tilde{u}}_i = \bm{u}_i / \|\bm{u}_i\|$, $\bm{\tilde{v}}_j = \bm{v}_j / \|\bm{v}_j\|$, and compute the mean vectors $\bm{\tilde{\mu}} = \sum_{i=1}^m \bm{\tilde{u}}_i / m$ and $\bm{\tilde{\nu}} = \sum_{j=1}^n \bm{\tilde{v}}_j / n$, the APD can be interpreted as a kind of distance between these mean vectors:$f_{\mathrm{APD}}(w) = 1 - \bm{\tilde{\mu}}^\top \bm{\tilde{\nu}}$.}: 
\begin{align*}
    f_{\mathrm{APD}}(w)= \frac{1}{mn}\sum_{i,j}(1-\cos(\bm u_i, \bm v_j)).
\end{align*}
Moreover, we used standard OT distance, which was defined as
\begin{align*}
    f_\mathrm{OT}(w) = \sum_{i,j} C_{ij}T_{ij},
\end{align*}
where $\bm{T}$ was the optimal solution for the standard OT problem~\eqref{eq:OT}. We also defined a metric based on LDR for each usage instance. Let $\mathrm{LDR}(s_i)$ denote the value of LDR for $s_i$. Following the definition of $f_\mathrm{SUS}$, we defined $f_\mathrm{LDR}$ as
\begin{align*}
    f_\mathrm{LDR}(w)  =  \left| \frac{1}{m}\sum_{i=1}^m \mathrm{LDR}(s_i)- \frac{1}{n}\sum_{j=1}^n \mathrm{LDR}(t_j)\right|.
\end{align*}

As a sense-based approach, we used WiDiD to estimate the normalized SFDs $\hat{P}$ and $\hat{Q}$ for the old and modern corpora, respectively. Using these estimated SFDs, we defined the two metrics $f_{\mathrm{WiDiD}}$ and $f_{\mathrm{APDP}}$. The former was defined by $f_{\mathrm{WiDiD}}(w) = \mathrm{JSD}(\hat P, \hat Q)$. The latter was defined as the APD between sense prototypes (APDP). APDP is calculated as follows: For each sense $k$, compute the mean embeddings $\bm{\mu}_k$ and $\bm{\nu}_k$ for the old and modern corpora, respectively. The APDP measures the APD between the sets of sense prototypes $\{\bm{\mu}_k\}_k$ and $\{\bm{\nu}_k\}_k$. Following \citet{periti-tahmasebi-2024-systematic}, we used the Canberra distance as the metric for APD.

\paragraph{Results.} 
Table~\ref{tab:appx-JSD-rank-corr} shows the Spearman rank correlation between the gold change score $f^*$ and the change score produced by each method. The form-based approach that directly uses embeddings achieves the highest performance. However, the change scores based on SUS demonstrate comparable performance, confirming that SUS effectively captures word-level semantic change.

\subsection{Quantifying the word-level change in semantic scope}
\label{appx:word-shift-signed}

\begin{table}[t]
\small
    \centering
    \begin{tabular}{ccc}
        \toprule
          Method &Approach& Spearman\\
        \midrule
         $g_{\mathrm{SUS}}$ & SUS-based & 0.55\\
         $g_1$ & SUS-based & 0.45\\
         $g_{\mathrm{vMF}}$ &form-based & \textbf{0.62}\\
         $g_\mathrm{LDR}$ &form-based&0.36\\
         $g_{\mathrm{WiDiD}}$ & sense-based &0.40 \\
        
        \bottomrule
    \end{tabular}
    \caption{Performance of methods for measuring word-level changes in semantic scope.}
    \label{tab:appx-ENT-rank-corr}
\end{table}

\paragraph{Gold change score.}
Following~\citet{DBLP:conf/acl/GiulianelliTF20}, the gold change score is defined as the entropy difference between $P^*$ and $Q^*$, i.e., $g^*(w) = g(P^*, Q^*)$.

\paragraph{Proposed change scores.}
For a target word, let the variances of SUS in the old and modern corpora be $ V_S = \mathrm{Var}(\{\mathrm{SUS}(s_i)\}_{i=1}^m)$ and $ V_T = \mathrm{Var}(\{\mathrm{SUS}(t_j)\}_{j=1}^n)$, respectively. In Section~\ref{sec:sus-based-word-level}, we only showed $g_\mathrm{SUS}$, whereas another metric was also designed:
\begin{align*}
    g_{\mathrm{SUS}}(w) &= \log\frac{V_T}{V_S},\\
    g_1 (w;\theta) 
    &= \sum_{i: \alpha_i<-\theta} \alpha_i +  \sum_{j: \beta_j>\theta} \beta_j.
\end{align*}
Here, $\alpha_i = \mathrm{SUS}(s_i)$ and $\beta_j = \mathrm{SUS}(t_j)$. The SUS value for each instance of a target word reflects the extent to which the usage in its sense has increased or decreased. Based on this, we aimed to quantify the change in semantic scope by summing the SUS values across all instances of the word. However, directly summing all SUS values results in a value of 0, i.e., $\sum_i \alpha_i + \sum_j \beta_j = 0$, due to the definition of SUS. To address this, we introduced a threshold $\theta$ to focus only on SUS values with a significant impact, calculating the sum of those values exceeding the threshold. The threshold $\theta$ is a hyperparameter. For details on the tuning process for $\lambda$ and $\theta$, refer to Appendix~\ref{appx:uot_detail}.

\paragraph{Baseline change scores.}
As a form-based approach, following \citet{nagata-etal-2023-variance}, we used a metric called coverage, defined as:  
\begin{align*}
    g_{\mathrm{vMF}}(w) = \log\frac{\kappa_S}{\kappa_T},
\end{align*} 
where $ \kappa_S$ and $ \kappa_T$ are known as concentration parameters, representing the spread of the distributions $ \{\bm{u}_i\}_i$ and $ \{\bm{v}_j\}_j$, respectively\footnote{These concentration parameters are derived from the norms $ \|\bm{\tilde{\mu}}\|$ and $ \|\bm{\tilde{\nu}}\|$, where $ \bm{\tilde{\mu}}$ and $ \bm{\tilde{\nu}}$ are the mean vectors of the normalized embeddings $ \{\bm{\tilde{u}}_i\}_i$ and $ \{\bm{\tilde{v}}_j\}_j$, respectively. The embeddings are assumed to follow a von Mises-Fisher (vMF) distribution, and the concentration parameters $ \kappa_S$ and $ \kappa_T$ represent the reciprocal of variance of these distributions.}.
We also defined a metric based on LDR for each usage instance. Let the variances of LDR in the old and modern corpora be $ U_S = \mathrm{Var}(\{\mathrm{LDR}(s_i)\}_{i=1}^m)$ and $ U_T = \mathrm{Var}(\{\mathrm{LDR}(t_j)\}_{j=1}^n)$, respectively. 
Following the definition of $g_\mathrm{SUS}$, we defined $g_\mathrm{LDR}$ as
\begin{align*}
    g_\mathrm{LDR}(w)  =  \log\frac{U_T}{U_S}.
\end{align*}

As a sense-based approach, we defined the metric using the entropy difference between the normalized SFDs $\hat{P}$ and $\hat{Q}$, which are estimated by WiDiD. This metric was expressed as $g_{\mathrm{WiDiD}}(w) = g(\hat{P}, \hat{Q})$.

\paragraph{Results.}
Table~\ref{tab:appx-ENT-rank-corr} shows the Spearman rank correlation between the gold change score $g^*$ and the change score calculated by each method. The form-based approach achieves the highest performance, indicating that using the values corresponding to the variance of embeddings in each time period effectively captures the broadening or narrowing of meaning. The change score based on SUS outperforms the sense-based method, demonstrating a certain level of validity.

\section{Experiments on Another Dataset}\label{appx:dwug_es}
\begin{table}[t]
\small
    \centering
    \begin{tabular}{ccc}
        \toprule
         Method & Approach& Spearman\\
        \midrule
         $f_{\mathrm{SUS}}$ & SUS-based & \textbf{0.61}\\
        $f_1$  & SUS-based& 0.60  \\
        $f_2$ & SUS-based & 0.60 \\
        $f_3$ & SUS-based &\textbf{0.61} \\
        $f_{\mathrm{OT}}$ &form-based & \textbf{0.61}\\
        $f_{\mathrm{APD}}$ &form-based & \textbf{0.61}\\
        $f_\mathrm{LDR}$&form-based  & 0.36\\
         $f_{\mathrm{WiDiD}}$ & sense-based  & 0.41 \\
        $f_{\mathrm{APDP}}$ & sense-based  &0.52 \\
        \bottomrule
    \end{tabular}
    \caption{Performance of methods for measuring the magnitude of word-level semantic change in the Spanish DWUG dataset.}
    \label{tab:appx-JSD-rank-corr-ES}
\end{table}

\begin{table}[t]
\small
    \centering
    \begin{tabular}{ccc}
        \toprule
          Method &Approach& Spearman\\
        \midrule
         $g_{\mathrm{SUS}}$ & SUS-based & 0.36\\
         $g_1$ & SUS-based & \textbf{0.45}\\
         $g_{\mathrm{vMF}}$ &form-based & 0.41\\
         $g_\mathrm{LDR}$ &form-based&0.31\\
         $g_{\mathrm{WiDiD}}$ & sense-based &0.33 \\
        \bottomrule
    \end{tabular}
    \caption{Performance of methods for measuring word-level changes in semantic scope in the Spanish DWUG dataset.}
    \label{tab:appx-ENT-rank-corr-ES}
\end{table}

Since the SUS-based metric evaluated in Section~\ref{sec:experiment_word-level} may be overfitting to the English DWUG dataset (DWUG EN), we conduct an evaluation experiment for word-level semantic change detection using the Spanish DWUG dataset~\cite{DBLP:conf/acl-lchange/Zamora-ReinaBS22} (DWUG ES) as an additional dataset to verify the effectiveness of SUS. The experimental setup including hyperparameter tuning follow the same procedure as in Section~\ref{sec:experiment_word-level}. The search scope for the hyperparameter is the same as in the experiment with DWUG EN.

\paragraph{Quantifying the magnitude of word-level semantic change.} 
Table~\ref{tab:appx-JSD-rank-corr-ES} shows the Spearman rank correlation between the gold change score $f^*$ and the change score produced by each method.
The SUS-based approach demonstrates performance comparable to the SOTA method $f_{\mathrm{APD}}$, 
suggesting that SUS is also effective in measuring the degree of semantic change in DWUG ES.

\paragraph{Quantifying the word-level change in semantic scope.} 
Table~\ref{tab:appx-ENT-rank-corr-ES} shows the Spearman rank correlation between the gold change score $g^*$ and the change score calculated by each method.
For $g_\mathrm{SUS}$, when compared to the baselines, the ranking of results is consistent with that in DWUG EN, and for $g_1$, it outperforms $g_{\mathrm{vMF}}$. 
This suggests that SUS is also effective in capturing changes in semantic scope in DWUG ES.

\end{document}